\documentclass{article}

\usepackage{microtype}
\usepackage{graphicx}
\usepackage{subcaption}

\usepackage[table]{xcolor} % 引入xcolor宏包，并开启table选项
\usepackage[most]{tcolorbox}
\usepackage{booktabs} % for professional tables
\usepackage{hyperref}

\usepackage[accepted]{icml2025}
\usepackage{amsmath}
\usepackage{amssymb}
\usepackage{mathtools}
\usepackage{amsthm}
\usepackage{xspace}
\usepackage{bm}
\usepackage{listings}
\usepackage{enumitem}
\usepackage{algorithm}
\usepackage{algorithmic}
\def\eqref#1{equation~\ref{#1}}
\def\1{\bm{1}}

\DeclareMathAlphabet{\mathsfit}{\encodingdefault}{\sfdefault}{m}{sl}
\SetMathAlphabet{\mathsfit}{bold}{\encodingdefault}{\sfdefault}{bx}{n}

\newcommand{\E}{\mathbb{E}}

\usepackage[capitalize,noabbrev]{cleveref}

\theoremstyle{plain}
\definecolor{theoremcolor}{rgb}{0.94, 0.94, 0.94}
\definecolor{examplecolor}{rgb}{1, 1, 1.0}
\newtheorem{proposition}{Proposition}
\newtheorem{theorem}{Theorem}

\theoremstyle{remark}
\newtheorem{remark}[theorem]{Remark}

\usepackage[textsize=tiny]{todonotes}

\usepackage{colortbl}
\usepackage[most]{tcolorbox}

\newcommand{\PP}{\textsc{Ppl}\xspace}
\newcommand{\SC}{\textsc{Sc}\xspace}
\newcommand{\Verb}{\textsc{Verb}\xspace}
\newcommand{\PC}{\textsc{Pc}\xspace}
\newcommand{\RP}{\textsc{Rp}\xspace}
\newcommand{\RPC}{\textsc{Rpc}\xspace}

\newcommand{\I}{\mathbb{I}\xspace}

\usepackage{multirow}
\newtcolorbox[auto counter, number freestyle={\noexpand\arabic{\tcbcounter}}]{promptbox}[2][]{%
    enhanced,
    colback=blue!5!white,
    colframe=black!75!white,
    title=Prompt~\thetcbcounter: #2,
    #1
}

\icmltitlerunning{Bridging Internal Probability and Self-Consistency for Effective and Efficient LLM Reasoning}

\begin{document}
\twocolumn[

\icmltitle{Bridging Internal Probability and Self-Consistency for \\ Effective and Efficient LLM Reasoning}

\begin{icmlauthorlist}
\icmlauthor{Zhi Zhou}{njucs}
\icmlauthor{Yuhao Tan}{njucs}
\icmlauthor{Zenan Li}{njucs}
\icmlauthor{Yuan Yao}{njucs}
\icmlauthor{Lan-Zhe Guo}{njucs,njusz}
\icmlauthor{Xiaoxing Ma}{njucs}
\icmlauthor{Yu-Feng Li}{njucs,njuai}
\end{icmlauthorlist}
\icmlaffiliation{njucs}{National Key Laboratory for Novel Software Technology, Nanjing University}
\icmlaffiliation{njuai}{School of Artificial Intelligence, Nanjing University}
\icmlaffiliation{njusz}{School of Intelligence Science and Technology, Nanjing University}
\icmlcorrespondingauthor{Xiaoxing Ma}{xxm@nju.edu.cn}
\icmlcorrespondingauthor{Yu-Feng Li}{liyf@nju.edu.cn}

\icmlkeywords{Large Language Model, Self Consistency, Math Reasoning, Code Generation}
\vskip 0.3in
]
\printAffiliationsAndNotice{}  % leave blank if no need to mention equal contribution

\begin{abstract}
    Recent advancements in large language models (LLMs) have demonstrated remarkable reasoning capabilities. However, single-shot inference often yields unreliable results for complex reasoning tasks, leading researchers to explore multiple reasoning paths through methods such as perplexity and self-consistency.
    In this paper, we present the first theoretical error decomposition analysis of these techniques, breaking down their error into estimation error and model error. Our analysis reveals a fundamental trade-off: perplexity methods suffer from substantial model error due to the absence of a proper consistency function, while self-consistency exhibits high estimation error due to a slow error convergence rate. 
    To overcome these limitations, we propose \emph{\textbf{R}easoning-Pruning \textbf{P}erplexity \textbf{C}onsistency} (\RPC). This approach combines \emph{Perplexity Consistency}, which seamlessly integrates LLM perplexity with self-consistency, and \emph{Reasoning Pruning}, which eliminates low-probability reasoning paths to effectively prevent the degeneration of estimation error reduction.
    Theoretical analysis demonstrates that \RPC not only accelerates the convergence rate of estimation error to an exponential level but also holds strong potential for further reducing model error. 
    Extensive empirical evaluations on seven benchmark datasets confirm that \RPC can significantly improve reasoning performance, sample efficiency, and confidence reliability.
\end{abstract}

\section{Introduction}

Recently, large language models (LLMs) have shown significant progress in various applications such as problem solving~\citep{LewkowyczADDMRS22, li24coc}, planning~\citep{ValmeekamMSK23plan, deng24plan}, and decision making~\citep{Ouyang023decision, SblendorioDCGPC24decision}, demonstrating their reasoning capabilities. 
Since single-shot inference is not always reliable, especially in complex reasoning tasks, one often requires the LLM to produce multiple reasoning paths, facilitating its reasoning performance.

When multiple reasoning paths for a given problem are available, the reasoning performance is determined by the confidence estimation for each result.
To achieve this, perplexity methods~\citep{chen1998evaluation, murugadoss2025evaluating} apply LLMs' internal probability to estimate the confidence of the reasoning path. 
Although the internal probability is quite accurate, the reasoning path confidence is highly insufficient to distinguish each answer, thereby greatly limiting the effectiveness of perplexity methods~\citep{chen24steering}.
In contrast,
self-consistency methods~\citep{wang2022self, chen2023universal} switch to establish the answer confidence using a pre-defined consistency function. 
However, the answer confidence cannot be directly derived from the internal probabilities of LLMs, necessitating the use of Monte-Carlo estimation, which significantly degrades the convergence rate~\citep{amad23adaptive, wang24dynamic, wang24make}.

\begin{figure}[t]
    \begin{center}
    \centerline{\includegraphics[width=\columnwidth]{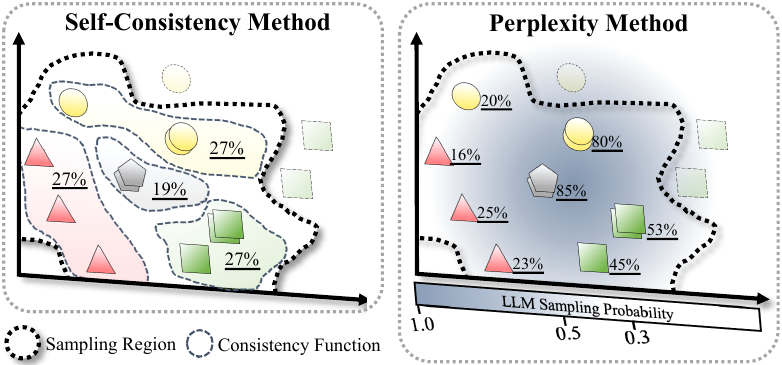}}
    \caption{
        The comparison between self-consistency and perplexity methods. Self-consistency uses Monte-Carlo estimation while perplexity methods directly use LLM prediction probability (i.e., computing probabilities using perplexity of LLM).}
    \label{fig:intro-comparison}
    \end{center}
    \vskip -0.4in
\end{figure}

To better understand the limitations of current methods and to guide the development of an effective and efficient LLM reasoning approach, we formulate the LLM reasoning problem and present a theoretical analysis that decomposes the reasoning error into two components: \emph{Estimation Error} and \emph{Model Error}. 
Self-consistency methods, which rely on the Monte-Carlo estimation, achieve only a linear estimation error reduction rate with respect to the sample size.
The linear convergence rate leads to the method requiring a large sampled budget.
For instance, implementing self-consistency with 64 samples on the MATH dataset using the GPT-4 API costs approximately \$2000~\citep{li24escape}, rendering it extremely expensive for both researchers and organizations.
As to perplexity methods, their estimation error converges exponentially as they use the internal probability of LLMs.
The exponential convergence rate ensures that perplexity methods can work well even in a very limited sample budget, while its final convergent result is far from satisfactory due to the high model error.
A comparison between self-consistency methods and perplexity methods is shown in Figure~\ref{fig:intro-comparison}.
This complementary between estimation error and model error raises a chance to further improve the LLM reasoning performance:
\emph{Can we design a method that achieves both fast estimation error convergence rate and low model error?}
To the best of our knowledge, the efforts in this aspect remain limited.

In this paper, we explore effectively and efficiently integrating the internal LLM probability into the self-consistency framework, allowing us to utilize an accurate probability for rapid estimation error reduction while maintaining low model error.
We name this confidence estimation approach \emph{Perplexity Consistency}.
Our theoretical study illustrates that perplexity consistency provides a tight integration and can indeed achieve the goal.
However, the reduction rate of perplexity consistency estimation error undesirably degenerates to a linear rate when the magnitude of the LLM's internal probability is low. 
To tackle this issue, we further introduce \emph{Reasoning Pruning} to automatically model the probability distribution for each reasoning problem and remove low-probability reasoning paths. 
Combining the perplexity consistency and the reasoning pruning, we propose \emph{\textbf{R}easoning-pruning \textbf{P}erplexity \textbf{C}onsistency} (\RPC).

Our theoretical and experimental results confirm the efficient and effective performance of \RPC. 
Specifically, on four mathematical reasoning datasets, \RPC successfully reduces the sampling budget by at least 50\% while achieving the same reasoning performance as self-consistency. 
Conversely, with an equal sampling budget, \RPC outperforms existing methods by 1.29\% on average.
Additionally, \RPC provides confidence estimates that align better with the ground truth compared to existing methods.

To summarize, the main contributions of the paper are:

(1) We formulate the LLM reasoning problem and offer a theoretical analysis that decomposes LLM reasoning performance into estimation error and model error. This analysis emphasizes the benefits of self-consistency while revealing its limitations when working with limited sampling budgets.

(2) Building on our theoretical framework, we introduce the \RPC, which integrates \emph{Perplexity Consistency} and \emph{Reasoning Pruning}. This approach utilizes precise LLM probabilities and eliminates low-probability reasoning paths to enhance reasoning performance.

(3) Our theoretical analysis shows that \emph{Perplexity Consistency} achieves an exponential error reduction rate in most cases, and \emph{Reasoning Pruning} effectively compensates for the remaining degeneration issues. 

(4) Through extensive experiments conducted on four mathematical reasoning and three code generation tasks, our proposed \RPC delivers promising results of improving both accuracy and confidence consistency.

\section{Problem and Analysis}
\label{sec:problem}

In this section, we start by outlining the problem formulation of LLM reasoning through sampling multiple reasoning paths.
Then, we provide a theoretical analysis that decomposes LLM reasoning performance into estimation error and model error. 
Finally, we present experimental results verifying our theoretical analysis. 
Our theoretical and empirical analysis motivates our follow-up method design. 

\subsection{Problem Formulation}

Given a reasoning problem $(x, y)$, where $x$ represents the input query, and $y$ represents the ground-truth answer. 
The LLM generates a reasoning path $\hat{t} = (t_1, \ldots, t_m)$ by sequentially sampling tokens according to the conditional probability distribution $p(t_i \,|\, x, t_{<i})$, where $m$ denotes the length of the reasoning path. 
The probability of generating the reasoning path $\hat{t}$ is defined as $p(\hat{t} \,|\, x)$, a.k.a the confidence of the reasoning path $\hat{t}$.
An answer extraction function $g(\cdot)$ maps the reasoning path to the final answer $\hat{y} = g(\hat{t})$, and the reasoning correctness is evaluated by the indicator function $\I[\hat{y} = y]$. 
We can extend the probability to the answer $\hat{y}$, i.e., the answer confidence, denoted as $p(\hat{y} \,|\, x)$.

The confidence essentially represents the probability that the reasoning path $\hat{t}$ or answer $\hat{y}$ is correct, which enables LLMs to select the most reliable solution among multiple candidates.
Nevertheless, enumerating all reasoning paths or answers is unfeasible; 
we have to estimate the LLM confidence based on finite $n$ sampled reasoning paths instead. 
Furthermore, 
to measure the reasoning performance of LLMs, we use the squared error of confidence estimation $\hat{p}(\hat{t}\,|\,x)$ to the reasoning path $\hat{t}$:
\begin{equation*}
\mathcal{E}(p) = \big(\hat{p}(\hat{t} \,|\, x) - \I[g(\hat{t}) = y] \big)^2.
\end{equation*}
If we can extend the confidence estimation to the answer $\hat{y}$, the squared error can be reformulated as
\begin{equation*}
\mathcal{E}(p) = \big(\hat{p}(\hat{y} \,|\, x) - \I[\hat{y} = y] \big)^2.
\end{equation*}
Below, we analyze two confidence estimation methods, i.e., self-consistency method~\citep{wang2022self} and perplexity method~\cite{huang2023look}. Specifically, the self-consistency method computes the answer confidence using Monte-Carlo estimation based on a consistency function $\mathbb{I}_C$, while the perplexity method directly computes the confidence of reasoning paths using internal LLM probabilities. 

\subsection{Theoretical Analysis}

To maximize the reasoning performance of LLMs, 
self-consistency methods (denoted as \SC)~\citep{xiong2023can, yadkori2024believe, becker2024cycles} often sample $n$ reasoning paths $\tilde{t}_1, \dots, \tilde{t}_n$, and then estimate the probability of each answer by 
\begin{equation*}
    \begin{aligned}
\hat{p}^{(\SC)}(\hat{y} \,|\, x) 
&= \frac{1}{n} \sum_{i=1}^n \I[\tilde{y}_i = \hat{y}], \quad \tilde{y}_i = g(\tilde{t}_i).
    \end{aligned}
\end{equation*}
Then, the reasoning error of the \SC method for a given problem $(x,y)$ can be computed by
\begin{equation*}
\begin{aligned}
\mathcal{E}(\hat{p}^{(\SC)}) &= \E_{\tilde{t}_i \sim p(t \,|\, x)}( \hat{p}^{(\SC)}(\hat{y} \,|\, x) - \mathbb{I}[\hat{y} = y] )^2 \\
&=\E_{\tilde{t}_i \sim p(t \,|\, x)}( \frac{1}{n} \sum_{i=1}^n \I[\tilde{y}_i = \hat{y}] - \mathbb{I}[\hat{y} = y] )^2.
\end{aligned}
\end{equation*}

To illustrate the key factors affecting the reasoning error, 
we provide an error decomposition in the following proposition. 

\begin{proposition}[\SC Reasoning Error Decomposition]
\label{prop:sc-reasoning-error-decomposition}
For any input $x$ with ground-truth answer $y$, let $\hat{p}^{(\SC)}(\hat{y} \,|\, x)$ denote the estimated probability of $\hat{y}$ by \SC.
Then, the reasoning error $\mathcal{E}(\hat{p}^{(\SC)})$ can be divided into two components: 
\begin{equation*}
    \begin{aligned}
        \mathcal{E}(\hat{p}^{(\SC)}) 
        = & \underbrace{\frac{1}{n} p(\hat{y} \,|\, x) (1- p(\hat{y}\,|\, x))}_{\text{Estimation Error}} \\
        & \qquad +  \underbrace{\big (p(\hat{y} \,|\, x) - \mathbb{I}[\hat{y} = y] \big )^2}_{\text{Model Error}}. 
    \end{aligned}
\end{equation*}
\end{proposition}

\begin{remark}
The detailed proof is provided in Appendix~\ref{app:props}.
The estimation error refers to the error caused by the finite sampling from the LLM probability, while the model error indicates the LLM's limited reasoning capability.
Note that the estimation error of \SC reduces to only the variance as the sampling is unbiased.
This proposition demonstrates that, aside from the \emph{model error}, which is determined by the LLM's inherent reasoning capabilities,
the reasoning error is bounded by the \emph{estimation error}.
Moreover, the estimation error reduction rate of the sample size is linear, resulting in a large error margin when the sampling is insufficient.
\end{remark}

To effectively offset the estimation error, we switch to analyze the reasoning error of perplexity methods (denoted as \PP).
In contrast to the \SC method that estimates the answer probability using the Monte-Carlo estimation, \PP directly utilizes the internal probability of LLMs $p(\tilde{t}_i \,|\, x)$ for the sampled reasoning path $\tilde{t}_i$. 
Therefore, for given the unique set of $n$ sampled reasoning paths $\mathcal{R} = \left \{ \tilde{t}_1, \ldots, \tilde{t}_n \right \}$, the estimated probability of each reasoning path $\hat{t}$ is
\begin{equation*}
\begin{aligned}
\hat{p}^{(\PP)}(\hat{t} \mid x) &= \left\{
\begin{array}{ll}
p(\tilde{t}_i \,|\, x), & \text{if} \quad \hat{t} = \tilde{t}_i \\
0, & \text{otherwise}
\end{array}
\right. \\
&= \sum_{\tilde{t} \in \mathcal{R}} \I\left [\hat{t} = \tilde{t} \, \right ] p(\tilde{t} \,|\, x).
\end{aligned}
\end{equation*}
Similarly, we also use the mean squared error to measure the reasoning performance of \PP:
\begin{equation*}
\begin{aligned}
& \mathcal{E}(\hat{p}^{(\PP)}) = \E_{\tilde{t}_i \sim p(t \,|\, x)}( \hat{p}^{(\PP)}(\hat{t} \,|\, x) - \mathbb{I}[\hat{y} = y] )^2 \\
& \qquad =\E_{\tilde{t}_i \sim p(t \,|\, x)}( \sum_{\tilde{t}\in \mathcal{R}} \I\left [\hat{t} = \tilde{t} \, \right ] p(\tilde{t} \,|\, x) - \mathbb{I}[g(\hat{t}) = y] )^2.
\end{aligned}
\end{equation*}
Now, we can obtain the following proposition.
\begin{proposition}[\PP Reasoning Error Decomposition]
\label{prop:ppl-reasoning-error-decomposition}
For any given input $x$ with ground-truth answer $y$, let $\hat{p}^{(\PP)}(\hat{t} \,|\, x)$ denote the estimated probability of $\hat{t}$ by \PP method.
Then, the reasoning error $\mathcal{E}(\hat{p}^{(\PP)})$ can be divided into two components: 
\begin{equation*}
    \begin{aligned}
        \mathcal{E}(\hat{p}^{(\PP)}) = 
        & \underbrace{(1 - p(\hat{t} \,|\, x)) ^ n {p}(\hat{t} \,|\, x) ( 2 \I[\hat{y}_i = y] - p(\hat{t} \,|\, x) ) }_{\text{Estimation Error}} \\
        & \qquad + \underbrace{\big ( p(\hat{t} \,|\, x) - \mathbb{I}[g(\hat{t}) = y] \big )^2}_{\text{Model Error}}. \\
    \end{aligned}
\end{equation*}
\end{proposition}

\begin{remark}
The detailed proof is provided in Appendix~\ref{app:props}.
Compared with \SC, the estimation error of \PP decreases exponentially, which is much faster. However, the model error of \PP is usually larger than that of \SC in practice. In Appendix~\ref{subsec:model-error-comparison-ideal}, we provide Proposition~\ref{prop:ideal-model-error-comparison} to demonstrate that \SC achieves a smaller model error than \PP in the ideal case, due to the advantages of the consistency function.
\end{remark}

\subsection{Empirical Observations}

\begin{figure}[t]
    \begin{center}
    \includegraphics[width=\columnwidth]{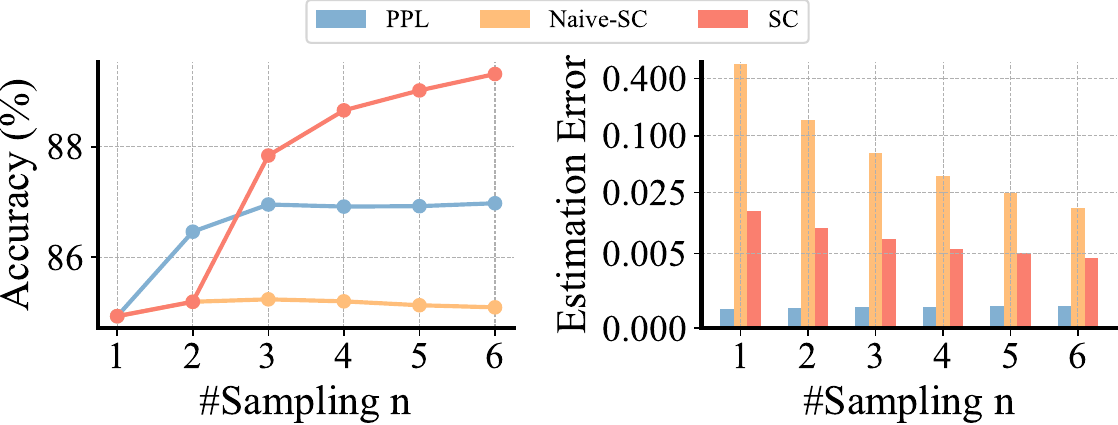}
    \caption{The performance of InternLM-MATH-Plus 7B model on GSM8K dataset. (1) Estimation error of \PP converges faster than \SC and Na\"{i}ve-\SC; (2) Proper consistency function is the key for \SC to achieve the small model error.}
    \label{fig:motivation-estimation-error}
    \end{center}
    \vskip -0.2in
\end{figure}

To confirm our theoretical results, we conduct some initial experiments on the GSM8K dataset using the InternLM-MATH-Plus 7B model. We limit the sample size $n$ from $1$ to $6$ and plot the accuracy curves and the estimation error in Figure~\ref{fig:motivation-estimation-error}.
Additionally, we include an ablative version called by Na\"{i}ve-\SC. 
Na\"{i}ve-\SC applies the Monte-Carlo estimation, which is consistent with \SC, 
but its consistency function is degraded to a Na\"{i}ve version to the reasoning path matching rather than the answer matching, which is consistent with \PP.
In other words, the reasoning error $\mathcal{E}(\hat{p}^{(\text{Na\"{i}ve-}\SC)})$ of Na\"{i}ve-\SC can be decomposed as
\begin{equation*}
    \begin{aligned}
        \underbrace{\frac{1}{n} p(\hat{t} \,|\, x) (1- p(\hat{t}\,|\, x))}_{\text{Estimation Error}}  +  \underbrace{\big (p(\hat{t} \,|\, x) - \mathbb{I}[g(\hat{t}) = y] \big )^2}_{\text{Model Error}}. 
    \end{aligned}
\end{equation*}

The derived results highlight the following two observations:

\textbf{(I) Estimation Error.}
The estimation errors of both \SC and \PP decrease as the sample size increases. 
However, the accuracy curves and estimation error illustrate that the \PP has a much faster convergence rate compared to \SC. 
Specifically, \PP reaches a stable result with $n=3$, while \SC cannot converge even for $n=6$.
Na\"{i}ve-\SC confirms this result, showing a lower convergent rate, since it uses the same Monte-Carlo estimation with \SC.

\textbf{(II) Model Error.}
\SC and \PP ultimately converge to different results. This is because their model errors are intrinsically different. \SC groups reasoning paths that yield the same answer through its consistency function, ensuring a higher accuracy of \SC. In contrast, \PP only estimates the probability of individual reasoning paths without considering answer-level consistency. Na\"{i}ve-\SC also supports this conclusion, converging to the worst results due to its lack of a proper consistency function.

{\bf Key Insights.} 
Our theoretical and empirical analyses point to the deficiencies and potential synergies of \SC and \PP. 
Although \SC achieves a lower model error due to the advantages of the consistency function, its estimation error is hindered by a slower convergence rate. In contrast, \PP exhibits a much faster convergence rate in estimation error using LLM internal probabilities, but this comes at the cost of a higher model error. This naturally raises a fundamental research question: \emph{Can we design a method fusing strengths of \SC and \PP: achieve both a rapid estimation error convergence rate and a low model error simultaneously?}

\begin{figure*}[t]
    \begin{center}
        \includegraphics[width=0.9\linewidth]{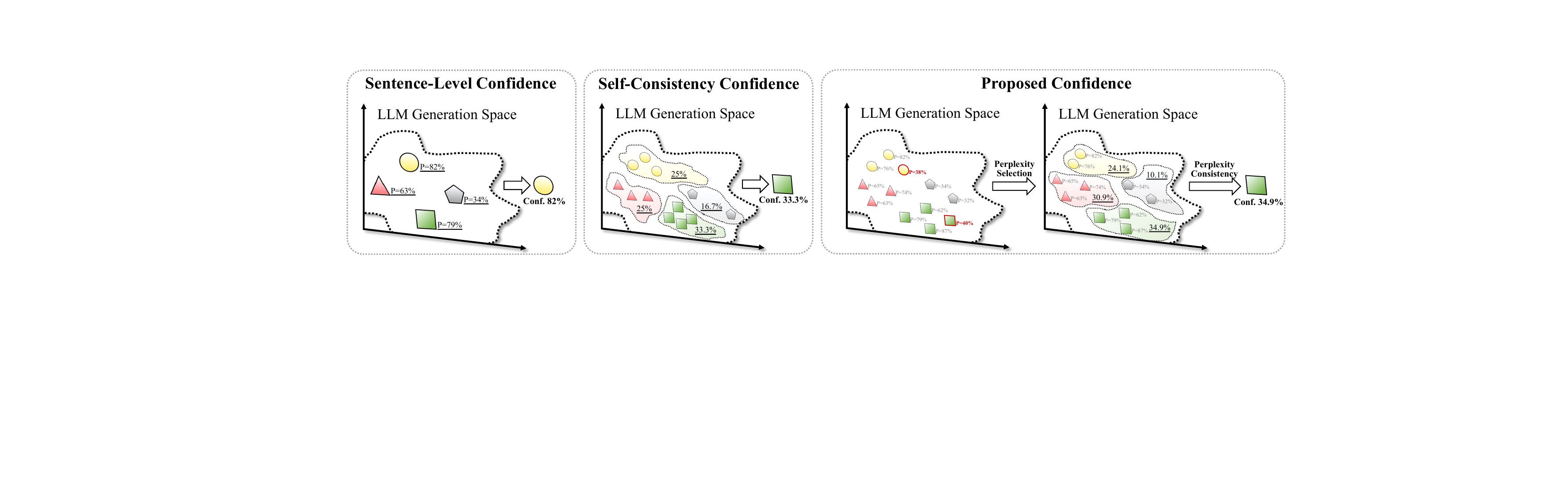}
        \caption{Illustration of the \RPC approach. The \emph{Reasoning Pruning} filters out low-probability answers, while the \emph{Perplexity Consistency} incorporates LLM probabilities into the self-consistency framework, resulting in faster convergence of estimation error.}
        \label{fig:framework}
    \end{center}
    \vskip -0.2in
\end{figure*}

\section{Methodology}
\label{sec:method}

Based on our theoretical and empirical analysis, we propose a new method called \emph{Reasoning-Pruning Perplexity Consistency} (\RPC). 
Specifically, we first integrate internal LLM probability into the self-consistency framework, paving the way for a confidence estimation function called \emph{Perplexity Consistency} (\PC). 
This function utilizes LLM probabilities to reduce estimation error more efficiently while maintaining a low model error.
Our further analysis guides the design of a new \emph{Reasoning Pruning} (\RP) module that addresses the limitations of \PC by filtering out reasoning paths with low probabilities.
\autoref{fig:framework} shows the overall framework.

\subsection{Perplexity Consistency}

To improve the efficiency of estimation error reduction, we propose \PC, which directly leverages the LLM's prediction probability like \PP, obtaining the benefit of an exponential convergent rate; 
and also applies the consistency function of \SC, to minimize the model error.
Formally, for the unique set of $n$ sampled reasoning paths $\mathcal{R} = \{\tilde{t}_1, \dots, \tilde{t}_n\}$, the estimated probability of the answer is
\begin{equation*}
\begin{aligned}
\hat{p}^{(\PC)}(\hat{y} \,|\, x) = \sum_{\tilde{t} \in \mathcal{R}} \I[g(\tilde{t}) = \hat{y}] p(\tilde{t} \,|\, x),
\end{aligned}
\end{equation*}
which calculates the cumulative probability of all unique reasoning paths whose answer is $\hat{y}$.
Therefore, the mean squared error of \PC is
\begin{equation*}
    \begin{aligned}
      \mathcal{E}(\hat{p}^{(\PC)}) = \E_{\tilde{t}_i \sim p(t \,|\, x)} \big[( \hat{p}^{(\PC)}(\hat{y} \,|\, x) - \I[\hat{y} = y] )^2 \big]. 
\end{aligned}
\end{equation*}
        
Now, we present the following theorem, which explores the reasoning error decomposition of \PC.

\begin{theorem}[\PC Reasoning Error Decomposition] \label{thm:thm1}
Assume that $k = |\{\tilde{t} \mid g(\tilde{t}) = \hat{y}\}|$ and define $\alpha := 1 - \frac{1}{k} p(\hat{y} \,|\, x)$. 
Then, the reasoning error $\mathcal{E}(\hat{p}^{(\PC)})$ of \PC can be divided into two components:
\begin{equation*}
    \begin{aligned}
      \mathcal{E}(\hat{p}^{(\PC)}) 
      = & \underbrace{ \alpha^n p(\hat{y} \,|\, x) \big(2\I[\hat{y}=y] - (1 + \alpha^n) p(\hat{y} \,|\, x) \big) }_{\text{Estimation Error}} \\
        & \qquad + \underbrace{\left( p(\hat{y} \,|\, x) - \I[\hat{y}_i = y] \right )^2}_{\text{Model Error}}. \\
    \end{aligned}
\end{equation*}
\end{theorem}
\begin{remark} 
The proof is presented in Appendix~\ref{app:thm1}. 
The theorem states that \PC successfully fuses the strengths of \PP and \SC: it achieves the same level of model error as \SC while ensuring the same convergence rate as \PP in the estimation error.
Particularly, the convergence rate can be computed as $\alpha^n p(\hat{y}\,|\, x) = (1 - \frac{1}{k} p(\hat{y} \,|\, x))^{n}p(\hat{y}\,|\, x)$. 
\end{remark}

The convergence rate is primarily influenced by the magnitude of $p(\hat{y} \,|\, x)$. In most scenarios, it remains exponential, facilitating rapid estimation error reduction.
However, when $p(\hat{y} \,|\, x) \to 0$ and $np(\hat{y} \,|\, x) \ll 1$, we only have $\alpha^n \to \frac{1}{1 + n p(\hat{y} \,|\, x)}$~\citep{kozma2021useful}, resulting in the convergence rate unexpectedly degenerating to a linear result.

\subsection{Reasoning Pruning}

Our analysis of the convergence rate in estimation error suggests some possibility for further improving the estimation error reduction efficiency.
Specifically, the rate degenerates when $p(\hat{y} \,|\, x)$ is too small.
Therefore, we propose directly pruning these low-probability answers instead of sampling, called by \emph{Reasoning Pruning} (\RP).

Reasoning pruning essentially aims to set $\hat{p}(\hat{y} \,|\, x) = 0$, i.e., when the cumulative probability of all its corresponding reasoning paths are very low, 
making the estimation error vanish. 
Although pruning low-probability answers can boost the efficiency of estimation error reduction, it also induces a level of model error. 
In other words, it may ignore some correct answers of low probability, thus implicitly degrading the performance of LLM reasoning.  

Ideally, if we know the probability $p(y \,|\, x)$ that the LLM can generate the correct answer, the optimal threshold for pruning should be $\tau = p(y \,|\, x)$. In this case, one can obtain not only an exponential convergence rate but also a significant reduction in model error, as all incorrect answers $\hat{y}$ satisfying $p(\hat{y} \,|\, x) < \tau$ are eliminated. However, to obtain this optimal error reduction, two challenges need to be resolved: (1) we only have an estimate of $p(\hat{y} \,|\, x)$ since the accurate value is unknown; and (2) we cannot know the ground-truth probability $p(y \,|\, x)$, making the threshold difficult to determine. 

For the first problem, we propose directly pruning reasoning paths instead of answers in \RP, with the following theorem. 
\begin{theorem} [Effectiveness of Reasoning Path Pruning]\label{thm:thm2}
Assume that the optimal threshold $\tau = p(y \,|\, x)$, and let $\hat{k} = |\{\tilde{t}_i \mid g(\tilde{t}_i) = \hat{y}, i=1,\dots,n\}|$, which refers to the size of samples whose answer is $\hat{y}$. 
Hence, \RP achieves the optimal error reduction with at least the probability 
\begin{equation*}
1 - \exp\Big(-{2\hat{k}k^2} (1 - \frac{\tau}{1 - \alpha})^2\Big).
\end{equation*}
\end{theorem}
\begin{remark}
The proof is included in Appendix~\ref{app:thm2}. 
The theorem provides a guarantee that \RP can achieve the optimal error reduction for each given problem $(x, y)$ at a high probability. 
Note that the optimal error reduction not only boosts the estimation error reduction efficiency but also effectively reduces the model error, thus improving the final reasoning capability of LLMs.
\end{remark}

The remaining problem is determining the optimal threshold for reasoning path removal. 
To achieve this goal, we develop an automated strategy. 
Inspired by open-set recognition~\citep{Bendale16openmax}, we model the probability distribution of $\Omega_1$ and $\Omega_2$ as a mixture of two Weibull distributions, representing high and low probability regions.

Elaborately, we define the PDF of mixture distribution as: 
\begin{equation*} \label{eq:weibull-mix}
    f(x) = w_1 f_{\text{W}}(x; k_1, \lambda_1) + w_2 f_{\text{W}}(x; k_2, \lambda_2), 
\end{equation*}
where the Weibull PDF~\citep{weibull1951statistical} is defined as $f_{\text{W}}(x; k, \lambda) = \frac{k}{\lambda}\left ( \frac{x}{\lambda}\right ) ^{k-1} \exp{\left ( -(\frac{x}{\lambda})^k\right )}$.
We use the maximum likelihood estimation methods to estimate the parameters, i.e., $(k_1, \lambda_1)$, $(k_2, \lambda_2)$, $w_1$, and $w_2$ on the probability distribution of all sampled reasoning paths for each reasoning problem. 
We assume that $\text{Weibull}(k_1, \lambda_1)$ is the high probability distribution and $\text{Weibull}(k_2, \lambda_2)$ is the low probability distribution. Then, the probability of reasoning path $\hat{t}$ being in the high probability distribution is derived by 
\begin{equation*}
    P_{\text{High}}(x) = \frac{w_1 f_{\text{W}}(x; k_1, \lambda_1)}{w_1 f_{\text{W}}(x; k_1, \lambda_1) + w_2 f_{\text{W}}(x; k_2, \lambda_2)}. 
    \label{eq:weibull-prob}
\end{equation*}
where $x$ is the value of LLM internal probability. 

Then, we remove sampled reasoning paths $\tilde{t}$ satisfying $P_{\text{High}}(\hat{p}(\tilde{t}\,|\,x)) < 0.5$, which more likely to be in the low probability distribution. Moreover, to ensure the algorithm's stability when $n$ is limited, we employ the Truncated Mean method~\citep{marazzi1999truncated}, retaining outputs with a probability greater than the overall mean. This prevents the removal of too many reasoning paths due to potential inaccurate estimation of the mixture distribution. 

Overall, we apply the \emph{Reasoning Pruning} to all sampled reasoning paths $\tilde{t}_1, \dots, \tilde{t}_n$ for removing low probability reasoning paths and then compute the confidence based on \emph{Perplexity Consistency}, forming our proposed \underline{\textbf{R}}easoning-pruning \underline{\textbf{P}}erplexity \underline{\textbf{C}}onsistency confidence (\RPC). The pseudo-code is shown in Algorithm~\ref{alg:rpc} in Appendix~\ref{sec:appendix-rpc}.

\section{Experiments}
\label{sec:exp}

In this section, we first conduct experiments to answer the following research questions:
\begin{enumerate}[leftmargin=2pt,itemsep=1pt,parsep=0pt,topsep=3pt]
\item[] \textbf{\underline{RQ1}: Efficiency.} How does \RPC reduce the number of samples required to achieve comparable performance through faster convergence?
\item[] \textbf{\underline{RQ2}: Efficacy.} How does \RPC improve reasoning performance compared to existing methods?
\item[] \textbf{\underline{RQ3}: Reliability.} How does \RPC enhance the reliability of confidence estimation compared to existing methods?
\end{enumerate}
Additional discussions are devoted to further demonstrating the effectiveness of \RPC. 
Due to space limitations, supplementary experimental results are included in Appendix~\ref{app:detailed-results}.

\begin{table*}[t]
    \centering
    \caption{Efficiency comparison of \emph{Perplexity Consistency} module (\PC) and \RPC. The table shows the minimum number of samples needed to exceed the best performance of \SC, with reduction rates in bold when sampling is reduced.}
    \label{tab:InternLM2-7B-Reduction}
    % \vskip 0.15in
    \begin{center}
    \begin{small}
    \begin{sc}
    \resizebox{\linewidth}{!}{
    \begin{tabular}{c|rrrrrrrr}
    \bottomrule
    \toprule
    \multirow{2}{*}{Method} & \multicolumn{2}{c}{\textbf{MATH}} & \multicolumn{2}{c}{\textbf{MathOdyssey}} & \multicolumn{2}{c}{\textbf{OlympiadBench}} & \multicolumn{2}{c}{\textbf{AIME}} \\
    \cmidrule(lr){2-9}  & Accuracy & \#Samplings & Accuracy & \#Samplings & Accuracy & \#Samplings & Accuracy & \#Samplings \\
    \midrule
     Best of \SC & 50.57 & 64   & 28.32 & 112    & 11.07 & 128    & 9.40  & 128       \\
    \midrule
    \PC      & 50.63 & 32      & 28.51 & 112    & 11.07 & 128    & 9.00  & 64        \\
    \rowcolor{gray!20} $\Delta$ & +0.06 & \textbf{-50.0\%} & +0.19 & -0.0\% & 0.00 & -0.0\% & 0.00  & \textbf{-50.0\%}  \\
    \hline
    \RPC     & 51.16 & 32      & 29.31 & 32      & 11.07 & 64      & 9.50  & 48      \\
    \rowcolor{gray!20} $\Delta$ & +0.59 & \textbf{-50.0\%} & +0.99 & \textbf{-71.4\%} & 0.00 & \textbf{-50.0\%} & +0.10 & \textbf{-62.5\%} \\
    \bottomrule
    \toprule
    \end{tabular}}
    \end{sc}
    \end{small}
    \end{center}
    % \vskip -0.1in
\end{table*}

\begin{figure*}[t]
% \vskip 0.2in
\begin{center}
    \begin{minipage}[t]{\textwidth}
        \centering
        \begin{subfigure}[t]{0.23\textwidth}
            \centering
    \includegraphics[width=\linewidth]{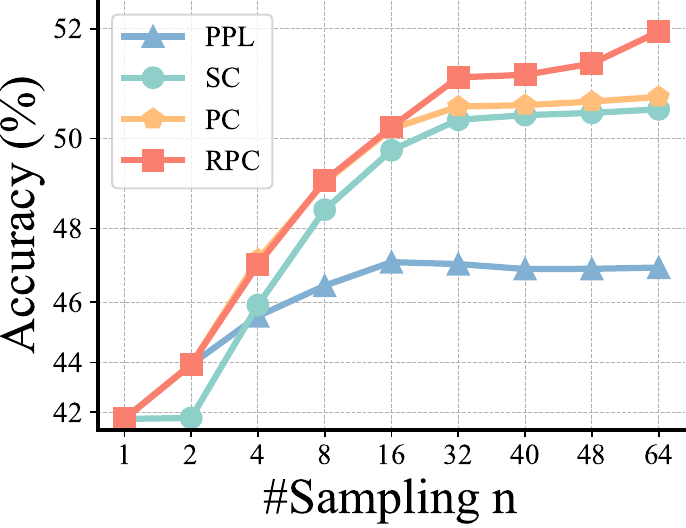}
            \vskip -0.3em
            \caption{MATH}
            \label{fig:MATH-Accuracy}
        \end{subfigure}
        \hfill
        \begin{subfigure}[t]{0.23\textwidth}
            \centering
            \includegraphics[width=\linewidth]{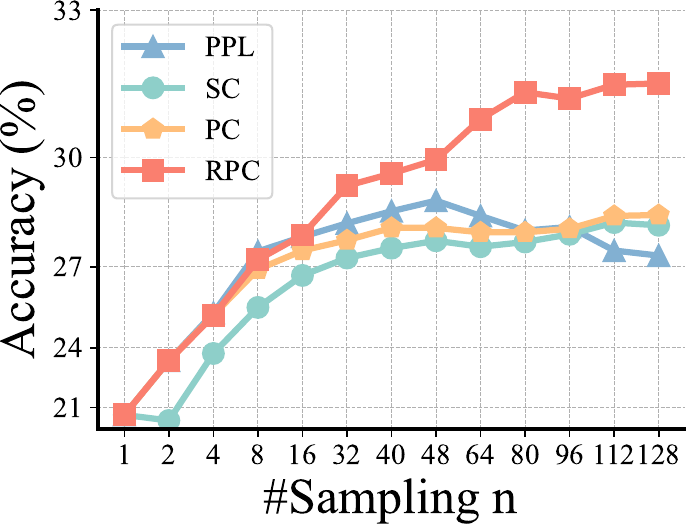}
            \vskip -0.3em
            \caption{MathOdyssey}
            \label{fig:MathOdyssey-Accuracy}
        \end{subfigure}
        \hfill
        \begin{subfigure}[t]{0.23\textwidth}
            \centering
            \includegraphics[width=\linewidth]{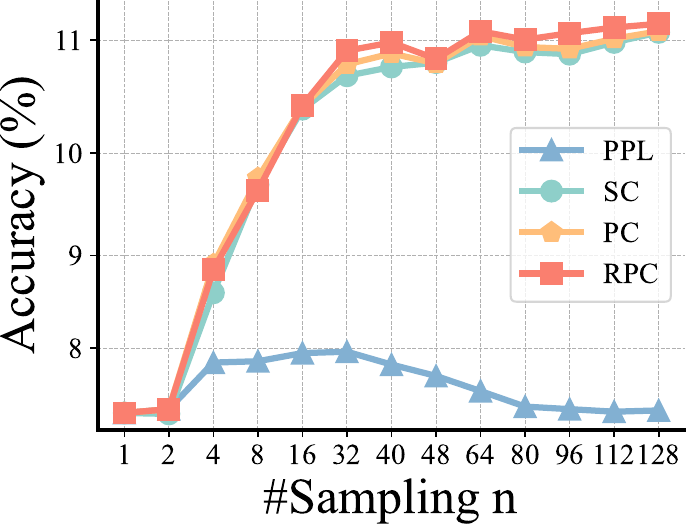}
            \vskip -0.3em
            \caption{OlympiadBench}
        \end{subfigure}
        \hfill
        \begin{subfigure}[t]{0.23\textwidth}
            \centering
            \includegraphics[width=\linewidth]{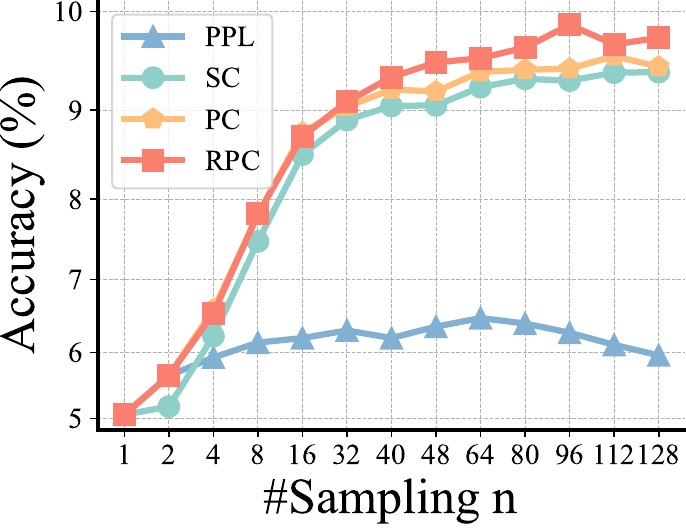}
            \vskip -0.3em
            \caption{AIME}
        \end{subfigure}
        \vskip -0.5em
        \caption{The accuracy of the InternLM-2-MATH-Plus 7B model on four math reasoning datasets with different sample sizes $n$. Our proposed \RPC achieves the best performance across all datasets, validating our theoretical analysis. We also show the performance of a single perplexity consistency module (denoted as \PC), which further supports our theoretical findings.}
        \label{fig:InternLM2-7B-Accuracy}
    \end{minipage}
\end{center}
\vskip -0.2in
\end{figure*}

\subsection{Experimental Setting}

In this section, we briefly introduce the comparison methods, datasets, and details of the implementation. 
The experimental settings can be found in Appendix~\ref{app:exp-details}.

\textbf{Comparison Methods.} 
We compare three types of LLM confidences: perplexity confidence~\citep{wang2022self} (\PP), self-consistency confidence~\citep{chen1998evaluation} (\SC), and verbalized confidence~\citep{tian2023just} (\Verb).
The verbalized confidence is computed based on the probability that the LLM outputs ``True'' versus ``False'' when asked an ``Is-True'' question. For code generation tasks, we extracted verbalized confidence scores from the model's numerical likelihood expressions by prompting the LLM.

\textbf{Datasets.} 
We introduce four popular benchmarks for math reasoning: MATH~\citep{hendrycks2021math}, MathOdyssey~\citep{Fang24Odyssey}, OlympiadBench~\citep{He24OlympiadBench}, and AIME~\citep{AIME}. 
As to code generation tasks, we evaluate each method on three benchmarks, i.e., HumanEval~\citep{Codex2021}, MBPP~\citep{MBPP2021}, and introductory-level problems of APPS~\citep{APPS2021}.
 
\textbf{Implementation Details.} 
For math reasoning tasks, we evaluate the InternLM2-Math-Plus models with 1.8B and 7B parameters~\citep{ying2024internlmmath}, as well as the DeepSeekMath-RL 7B model~\citep{shao24deepseekmath}. The consistency function $\mathbb{I}_C$ is answer comparison. For code generation tasks, we evaluate the Deepseek-Coder 33B model. The consistency function $\mathbb{I}_C$ is constructed based on semantic equivalence~\citep{SemanticEquiv2021} by clustering code based on given test cases. We set the sample size to $n=128$ for the MathOdyssey, OlympiadBench, and AIME datasets and $n=64$ for the MATH dataset by default. Each experiment is repeated 10 times with different random seeds, and the average performance is reported. All experiments were conducted on Linux servers with A800 and H800 GPUs.

\begin{table*}[t]
    \centering
    \caption{Performance Comparison using InternLM-2-MATH-Plus 7B model measured by accuracy and excepted calibration error metrics. The best performance is highlighted in \textbf{bold}. The results show that our \RPC outperforms existing methods in majority of cases.}
    \label{tab:InternLM2-7B-Performance}
    % \vskip -0.5in
    \vspace{-1em}
    \begin{center}
    \begin{small}
    \begin{sc}
    \resizebox{\linewidth}{!}{
    \begin{tabular}{l|cccccccc|cc}
    \bottomrule
    \toprule
    \multirow{2}{*}{Method} & \multicolumn{2}{c}{\textbf{MATH}} & \multicolumn{2}{c}{\textbf{MathOdyssey}} & \multicolumn{2}{c}{\textbf{OlympiadBench}} & \multicolumn{2}{c}{\textbf{AIME}} & \multicolumn{2}{c}{\textbf{Average}} \\
    \cmidrule(lr){2-11}     & Accuracy($\uparrow$) & ECE($\downarrow$) & Accuracy($\uparrow$) & ECE($\downarrow$) & Accuracy($\uparrow$) & ECE($\downarrow$) & Accuracy($\uparrow$) & ECE($\downarrow$) & Acc.($\uparrow$) & ECE($\downarrow$) \\
    \midrule
    \PP & 46.99 $\pm$ 0.20 & 48.99 $\pm$ 0.19 & 27.35 $\pm$ 1.22 & 67.70 $\pm$ 1.22 & ~~7.27 $\pm$ 0.36 & 86.90 $\pm$ 0.35 & 5.96 $\pm$ 0.48 & 88.98 $\pm$ 0.49 & 21.90  & 73.14 \\
    \Verb & 26.14 $\pm$ 0.25 & 47.46 $\pm$ 0.07 & 10.06 $\pm$ 0.61 & 69.92 $\pm$ 0.88 & ~~3.68 $\pm$ 0.16 & 84.68 $\pm$ 0.25 & 3.17 $\pm$ 0.17 & 86.29 $\pm$ 0.20 & 10.76  & 72.09  \\
    \SC & 50.57 $\pm$ 0.17 & ~~6.71 $\pm$ 0.18 & 28.25 $\pm$ 0.60 & 12.23 $\pm$ 0.54 & 11.07 $\pm$ 0.15 & 20.20 $\pm$ 0.16 & 9.40 $\pm$ 0.21 & \textbf{14.35 $\pm$ 0.23} & 24.82  & 13.37 \\
    \hline
    \rowcolor{gray!20} \RPC & \textbf{51.95 $\pm$ 0.15} & ~~\textbf{6.41 $\pm$ 0.18} & \textbf{31.62 $\pm$ 0.75} & ~~\textbf{9.87 $\pm$ 0.73} & \textbf{11.14 $\pm$ 0.15} & \textbf{18.86 $\pm$ 0.18} & \textbf{9.74 $\pm$ 0.23} & 14.32 $\pm$ 0.21 & \textbf{26.11} & \textbf{12.37} \\
    \bottomrule
    \toprule
    \end{tabular}}
    \end{sc}
    \end{small}
    \end{center}
    \vskip -0.1in
\end{table*}

\begin{figure}[t]
    % \vskip 0.1in
    \begin{center}
        \begin{minipage}[t]{\linewidth}
            \centering
            \begin{subfigure}[t]{0.48\linewidth}
                \centering
                \includegraphics[width=\linewidth]{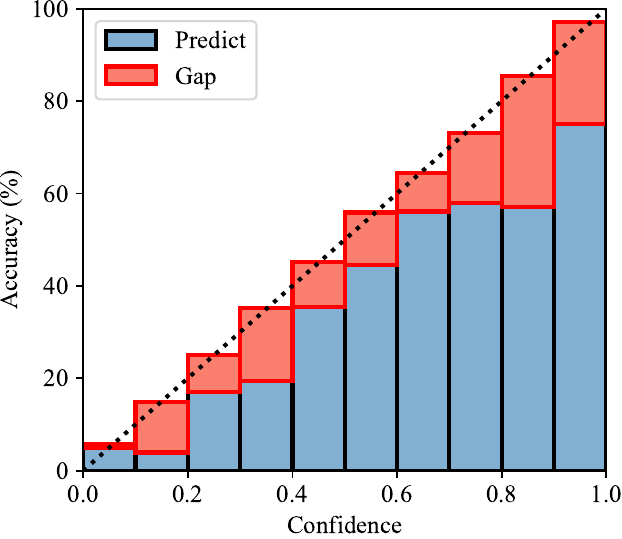}
                \vskip -0.3em
                \caption{\SC}
            \end{subfigure}
            \hfill
            \begin{subfigure}[t]{0.48\linewidth}
                \centering
                \includegraphics[width=\linewidth]{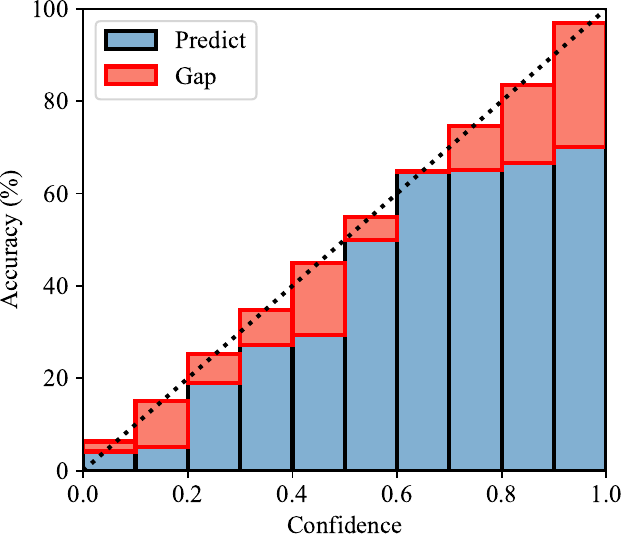}
                \vskip -0.3em
                \caption{\RPC}
            \end{subfigure}
            \vskip -0.5em
            \caption{The reliability diagrams of \SC and \RPC on MathOdyssey dataset using InternLM-2-MATH-Plus 7B model.}
            \label{fig:InternLM2-7B-Reliability}
        \end{minipage}
    \end{center}
    \vskip -0.3in
\end{figure}

\begin{table}[t]
    \centering
    \caption{Performance Comparison of different models and different parameter scales. The accuracy is reported as the mean and stdev. The best performance is highlighted in \textbf{bold}. The results show that our \RPC outperforms existing methods in major cases.}
    \label{tab:model-performance}
    \vspace{-1em}
    \begin{center}
    \begin{small}
    \begin{sc}
    \resizebox{\linewidth}{!}{
    \begin{tabular}{l|cccc}
    \bottomrule
    \toprule
    \multirow{2}{*}{Method} & \multicolumn{4}{c}{\textbf{ InternLM2-Math-Plus 1.8B}} \\
    \cmidrule(lr){2-5}      & MATH & MathOdyssey & OlympiadBench & AIME \\
    \midrule
    \PP & 33.24 $\pm$ 0.24 & \textbf{16.56 $\pm$ 0.88} & 3.08 $\pm$ 0.20 & 1.66 $\pm$ 0.15 \\
    \Verb & ~~7.21 $\pm$ 0.17 & ~~2.81 $\pm$ 0.26 & 0.77 $\pm$ 0.06 & 0.26 $\pm$ 0.05 \\
    \SC & 36.48 $\pm$ 0.15 & 14.52 $\pm$ 0.46 & 5.99 $\pm$ 0.17 & 2.66 $\pm$ 0.20 \\
    \hline
    \rowcolor{gray!20} \RPC & \textbf{37.88 $\pm$ 0.16} & 16.35 $\pm$ 0.61 & \textbf{6.52 $\pm$ 0.24} & \textbf{3.26 $\pm$ 0.24} \\
    \bottomrule
    \toprule
    \multirow{2}{*}{Method} & \multicolumn{4}{c}{\textbf{DeepSeekMath-RL 7B}} \\
    \cmidrule(lr){2-5}      & MATH & MathOdyssey & OlympiadBench & AIME \\
    \midrule
    \PP & 42.51 $\pm$ 0.23 & 22.34 $\pm$ 1.00 & ~~5.90 $\pm$ 0.31 & 3.37 $\pm$ 0.46 \\
    \Verb & 14.29 $\pm$ 0.23 & ~~2.55 $\pm$ 0.24 & ~~2.36 $\pm$ 0.15 & 1.91 $\pm$ 0.12 \\
    \SC & 53.33 $\pm$ 0.09 & 36.68 $\pm$ 0.65 & 11.29 $\pm$ 0.17 & 9.42 $\pm$ 0.23 \\
    \hline
    \rowcolor{gray!20} \RPC & \textbf{53.37 $\pm$ 0.11} & \textbf{37.25 $\pm$ 0.69} & \textbf{11.30 $\pm$ 0.11} & \textbf{9.52 $\pm$ 0.31} \\
    \bottomrule
    \toprule
    \end{tabular}}
    \end{sc}
    \end{small}
    \end{center}
    \vskip -0.1in
\end{table}
    
\subsection{Empirical Results}

\textbf{\underline{RQ1}: Efficiency.} How does \RPC reduce the number of samples required to achieve comparable performance through faster convergence?

We evaluate our proposed \RPC against the standard self-consistency method using four mathematical benchmark datasets with the InternLM-2-MATH-Plus 7B model. For the MATH dataset, we set the reasoning path size to 64, while we set the number of reasoning paths to 128 for the other datasets with \SC. We then record the best performance and minimum sampling requirements for \SC. For both \RPC and our \emph{Perplexity Consistency} module (denoted as \PC), we report the minimum number of samples needed to match or exceed the performance of the \SC in \autoref{tab:InternLM2-7B-Reduction}.

The results of \PC indicate improved convergence rates compared to \SC in several cases, while maintaining similar rates in others. These findings support the rapid convergence and degeneration issues of \PC in Theorem~\ref{thm:thm1}. \RPC shows significant efficiency improvements by requiring fewer samples to achieve comparable performance relative to \SC. Moreover, the degeneration issues observed in \PC are effectively addressed in \RPC, validating both the effectiveness of the \emph{Reasoning Pruning} module and our Theorem~\ref{thm:thm2}.

\textbf{\underline{RQ2}: Efficacy.} How does \RPC improve reasoning performance compared to existing methods?

We evaluate the performance of \PC and \RPC in \autoref{fig:InternLM2-7B-Accuracy} across various sample budgets. The results demonstrate that \RPC achieves better performance than both \PP (which relies on internal LLM probabilities) and \SC (which employs Monte Carlo sampling). The detailed accuracy results, including mean and standard deviation in \autoref{tab:InternLM2-7B-Performance} support these findings.

We also analyze the performance of \PC separately. The results indicate that \PC has a faster convergence rate than \SC, which aligns with Theorem~\ref{thm:thm1}. The significant performance gains from \PC to \RPC, as shown in \autoref{fig:MATH-Accuracy} and \autoref{fig:MathOdyssey-Accuracy}, validate the effectiveness of the \emph{Reasoning Pruning} module. This suggests that \emph{Reasoning Pruning} helps reduce model errors when the LLM exhibits good alignment by eliminating incorrect reasoning paths that carry low LLM probability scores.

\begin{figure}[t]
    \vskip 0.1in
    \begin{center}
    \centerline{\includegraphics[width=\columnwidth]{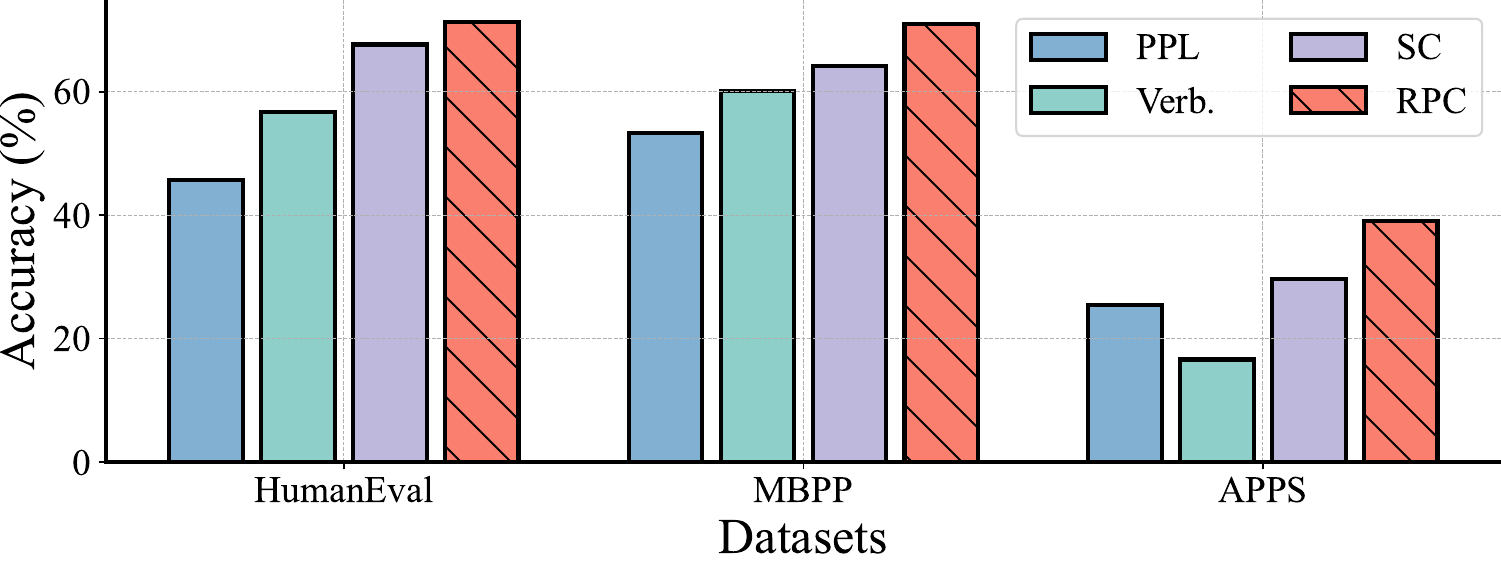}}
    \caption{Performance on code generation tasks using Deepseek-Coder 33B model. The results show that our \RPC achieves the best performance.}
    \label{fig:code-accuracy}
    \end{center}
    \vskip -0.3in
\end{figure}

\textbf{\underline{RQ3}: Reliability.} How does \RPC enhance the reliability of confidence estimation compared to existing methods?

To evaluate the reliability of confidence estimation, we analyze the ECE results of \RPC and comparison methods in \autoref{tab:InternLM2-7B-Performance}. ECE measures the difference between predicted probabilities and empirical accuracy, as directly computing estimation error using ground-truth probabilities is virtually impractical. The results demonstrate that our \RPC approach achieves lower ECE values and higher accuracy compared to \PP and \SC, indicating more reliable confidence estimation. We visualize this improvement through reliability diagrams comparing \SC and \RPC in \autoref{fig:InternLM2-7B-Reliability} on MathOdyssey, which clearly shows the reduced calibration error of \RPC.

\subsection{Further Discussion}

\textbf{Results on Code Generation Tasks.}
To investigate whether our proposed approaches can generalize to other reasoning tasks, such as code generation tasks, we evaluate \RPC and comparison methods on three code generation benchmarks, as illustrated in \autoref{fig:code-accuracy}. The results show that \RPC achieves the highest accuracy across all datasets, demonstrating its effectiveness in reasoning tasks beyond mathematics. 

\textbf{Results Across Model Scales and Architectures.}
To evaluate the generalization ability of our approaches across different model scales and architectures, we conducted additional experiments using InternLM2-Math-Plus 1.8B and DeepSeek-Math 7B models. The results in \autoref{tab:model-performance} show that \RPC consistently outperforms existing methods, which is consistent with results in \autoref{tab:InternLM2-7B-Performance}.

\section{Related Work}

This paper is related to the two research topics, i.e., LLM Reasoning Boosting and LLM Confidence Estimation.

\textbf{LLM Reasoning Boosting.}

Recent research has developed various methods to improve LLM reasoning capabilities. CoT~\citep{kojima22large} proposes the ``Let's think step by step'' prompt to guide LLMs in generating structured solutions. For complex problems, Least-to-Most~\citep{zhou23least} introduces a decomposition strategy that breaks down challenges into manageable sub-problems. Few-shot methods~\citep{wei22cot,fu23complexity,zhang23automatic} leverage carefully selected examples to improve reasoning performance. To enable more comprehensive reasoning, search-based methods~\citep{guan2025rstar} integrate Monte Carlo Tree Search (MCTS). Recent advancements have further enhanced MCTS by incorporating reward models~\citep{zhang2024restmcts, park24ensembling}. To directly optimize reasoning abilities, researchers have explored fine-tuning approaches~\citep{yu24metamath, li24mugglemath, li2024neurosymbolic,ying2024internlmmath} using specialized datasets and reinforcement learning techniques~\citep{shao24deepseekmath,luo25wizardmath}.

While these methods focus on generating accurate reasoning paths, our \RPC can build upon them by utilizing multiple sampling strategies, enabling better reasoning performance.

\textbf{LLM Confidence Estimation.}
The confidence estimation for LLM can be categorized into three types: (1) perplexity confidence, (2) verbalized confidence, and (3) self-consistency confidence. Perplexity confidence~\citep{huang2023look,duan2024shifting} utilizes the geometric mean of LLM prediction probabilities (i.e., perplexity~\citep{chen1998evaluation, blei03LDA}) to evaluate model adherence~\citep{murugadoss2025evaluating} and prompt quality~\citep{yao2024learning}. Verbalized confidence~\citep{kadavath2022language, xiong2023can, tian2023just} directly asks LLMs to express their confidence through various approaches, such as multi-agent deliberation~\citep{yang2024confidence}, multi-step evaluation~\citep{xiong2023can}, top-k ranking~\citep{tian2023just}, few-shot prompting~\citep{liu2023calibrating}, and reflection~\citep{dhuliawala2023chain, zhao2024fact}. Self-consistency confidence~\citep{wang2022self, chen2023universal, cheng2024relic} measures the agreement among multiple generated answers to improve reasoning performance, with recent work~\citep{xiong2023can, yadkori2024believe, becker2024cycles} further developing this approach as a confidence metric. Recent studies recognize its computational issues and propose early stopping~\citep{li24escape} and dynamic sampling methods~\citep{wang24make, wang24dynamic, amad23adaptive} to improve efficiency. 

Our proposed \RPC integrates LLM probability with a self-consistency framework, allowing perplexity and verbalized confidence to be used. \RPC achieves enhanced confidence estimation with fixed reasoning paths, showing a complement to existing self-consistency methods.

\section{Conclusion} \label{sec:concl}

In this paper, we address a foundational challenge in LLM reasoning: determining the most reliable answer from multiple reasoning paths by measuring LLM confidence. 
We present a theoretical framework to decompose the reasoning error into estimation error and model error, revealing that perplexity methods suffer from substantial model error due to lacking consistency function while self-consistency suffers from high estimation error because of a slow error convergence rate. 
To tackle this limitation, we introduce \emph{\textbf{R}easoning-pruning \textbf{P}erplexity \textbf{C}onsistency} (\RPC), a confidence estimation method with two key components: 
\emph{Perplexity Consistency} utilizes internal LLM probabilities to achieve faster estimation error convergence in major cases. 
\emph{Reasoning Pruning} prunes low-probability reasoning paths to prevent the remaining degeneration cases. 
Our theoretical analysis and extensive experiments demonstrate that \RPC achieves superior error convergence rates and reasoning performance compared to existing methods.

\textbf{Limitations and Future Work.} One limitation of this work is that we have only taken an initial step toward improving confidence estimation for self-consistency. The two components of \RPC can be further enhanced: \emph{Perplexity Consistency} could incorporate additional probability metrics from LLM outputs, while \emph{Reasoning Pruning} could be extended with more sophisticated pruning strategies, such as analyzing the probability distribution of each candidate answer. We believe our initial approach and theoretical analysis can guide future research in this promising direction.

\section*{Impact Statement}
This work advances the efficiency and effectiveness of LLM reasoning with multiple reasoning paths.  
Our method can benefit various applications requiring reliable artificial intelligence reasoning, such as mathematical problem-solving and code generation. There are many potential societal consequences of our work, none of which we feel must be specifically highlighted here.

\bibliography{ref}
\bibliographystyle{icml2025}

%%%%%%%%%%%%%%%%%%%%%%%%%%%%%%%%%%%%%%%%%%%%%%%%%%%%%%%%%%%%%%%%%%%%%%%%%%%%%%%
%%%%%%%%%%%%%%%%%%%%%%%%%%%%%%%%%%%%%%%%%%%%%%%%%%%%%%%%%%%%%%%%%%%%%%%%%%%%%%%
% APPENDIX
%%%%%%%%%%%%%%%%%%%%%%%%%%%%%%%%%%%%%%%%%%%%%%%%%%%%%%%%%%%%%%%%%%%%%%%%%%%%%%%
%%%%%%%%%%%%%%%%%%%%%%%%%%%%%%%%%%%%%%%%%%%%%%%%%%%%%%%%%%%%%%%%%%%%%%%%%%%%%%%
\newpage
\appendix
\onecolumn

\section{Theoretical Results}

\subsection{Proof of Proposition~\ref{prop:sc-reasoning-error-decomposition} and Proposition~\ref{prop:ppl-reasoning-error-decomposition}} \label{app:props}
\begin{proof}
  (\SC) First, we denote the sampling probability distribution of the LLM as $p(\hat{y} \,|\, x)$, 
  and the confidence function as $\hat{p}^{(\SC)}(\hat{y} \,|\, x) = \frac{1}{n}  \sum_{i=1}^n \mathbb{I}[\tilde{y}_i = \hat{y}_i] $, where $\tilde{y}_1, \dots, \tilde{y}_n$ are sampled on the distribution $p(\hat{y} \,|\, x)$. 
  Apply the error decomposition, we have
  \begin{equation*}
  \begin{aligned}
    \mathcal{E}(\hat{p}^{(\SC)}) & =  \E_{\tilde{y}_i \sim {p}(\tilde{y}_i \,|\, x)} \left[( \hat{p}^{(\SC)}(\hat{y} \,|\, x) - \I[\hat{y} = y])^2 \right]  \\
    & =  \E_{\tilde{y}_i \sim{p}(\tilde{y}_i \,|\, x)} \big[(\frac{1}{n}  \sum_{i=1}^n \mathbb{I}[\tilde{y}_i = \hat{y}_i] - \I[\hat{y} = y])^2 \big]  \\
    & = \E_{\tilde{y}_i \sim {p}(\tilde{y}_i \,|\, x)} \big[ ( \frac{1}{n}  \sum_{i=1}^n \mathbb{I}[\tilde{y}_i = \hat{y}_i] - p(\hat{y} \,|\, x) + p(\hat{y} \,|\, x) - \I[\hat{y}_i = y]  )^2 \big] \\
    & = \frac{1}{n} {p}(\tilde{y}_i \,|\, x) (1-{p}(\tilde{y}_i \,|\, x)) + \left( p(\hat{y} \,|\, x) - \I[\hat{y}_i = y] \right )^2.
  \end{aligned}
  \end{equation*}
   (\PP) Another way is to use the confidence function to build the sampling probability distribution, 
    i.e., 
    \begin{equation*}
    \hat{p}^{(\PP)}(\hat{t} \,|\, x) = \left\{
        \begin{array}{ll}
            p(\hat{t} \,|\, x), & \text{if}~ \text{there is}~\tilde{t}_i = \hat{t} \\
            0, & \text{otherwise}
        \end{array}
    \right. 
    = \sum_{i=1}^n \I(\tilde{t}_i = \hat{t})p(\hat{t} \,|\, x).
    \end{equation*}
    Now, we have
    \begin{equation*}
    \begin{aligned}
        & \E_{\tilde{y}_i \sim {p}(\tilde{t}_i \,|\, x)} [ \hat{p}^{(\PP)}(\hat{t} \,|\, x) - p(\hat{t} \,|\, x)]  = -(1 - p(\hat{t} \,|\, x)) ^ n p(\hat{t} \,|\, x) \\
        & \E_{\tilde{y}_i \sim {p}(\tilde{t}_i \,|\, x)} [ (\hat{p}^{(\PP)}(\hat{t} \,|\, x) - p(\hat{t} \,|\, x))^2] = (1 - p(\hat{t} \,|\, x)) ^ n p(\hat{t} \,|\, x)^2 .
    \end{aligned}
    \end{equation*}
    Hence,
    \begin{equation*}
        \begin{aligned}
          \mathcal{E}(\hat{p}^{(\PP)}) &= \E_{\tilde{t}_i \sim p(t \,|\, x)} \big[( \hat{p}^{(\PP)}(\hat{t} \,|\, x) - \I[\hat{y} = y] )^2 \big] \\
	& =\E_{\tilde{t}_i \sim p(t \,|\, x)}\big[(\hat{p}^{(\PP)}(\hat{t} \,|\, x) - p(\hat{t} \,|\, x) + p(\hat{t} \,|\, x) - \I[g(\hat{t}) = y] )^2 \big] \\
	& = - (1 - p(\hat{t} \,|\, x)) ^ n p(\hat{t} \,|\, x)^2 + 2 (1 - p(\hat{t} \,|\, x)) ^ n p(\hat{t} \,|\, x) \I[g(\hat{t}) = y] + (p(\hat{t} \,|\, x) - \I[g(\hat{t}) = y] )^2 \\
	& = (1 - p(\hat{t} \,|\, x)) ^ n p(\hat{t} \,|\, x) (2\I[g(\hat{t}) = y] - p(\hat{t} \,|\, x)) + (p(\hat{t} \,|\, x) - \I[g(\hat{t}) = y] )^2.
        \end{aligned}
        \end{equation*}
Hence, we complete the proof.
\end{proof}

\subsection{Model Error Comparison in Ideal Case}
\label{subsec:model-error-comparison-ideal}

\begin{proposition}[Comparison of Model Errors]
    Consider a setting with infinite sampling of reasoning paths ($n \to \infty$) where each incorrect reasoning path leads to a unique answer, that is, $g(\tilde{t}_i) \neq g(\tilde{t}_j)$ for any $i \neq j$ where $g(\tilde{t}_i) \neq y$ and $g(\tilde{t}_j) \neq y$. The model error of self-consistency ($\mathcal{E}_{\text{Model}}^{(\SC)}$) and perplexity ($\mathcal{E}_{\text{Model}}^{(\PP)}$) satisfy:
    \begin{equation}
        \mathcal{E}_{\text{Model}}^{(\SC)} \leq \mathcal{E}_{\text{Model}}^{(\PP)}
    \end{equation}
    The inequality is strict when the consistency function identifies equivalent correct reasoning paths more than once.
    \label{prop:ideal-model-error-comparison}
\end{proposition}

\begin{remark}
    Although the assumptions in Proposition~\ref{prop:ideal-model-error-comparison} are idealized, this special case demonstrates that the consistency function in self-consistency achieves better model error than the perplexity method. 
    In practice, the perplexity method always gives larger model error compared to the self-consistency method, as it does not leverage the consistency function to analyze the structural properties of specific reasoning problems.
    The proof is presented as follows.
\end{remark}

\begin{proof}
    We first recall the definitions of the model error for self-consistency and perplexity:
    \begin{equation}
        \begin{cases}
            \mathcal{E}_{\text{Model}}^{(\SC)} &= \sum_{\hat{y} \in \{g(\tilde{t}_i) \, | \, i = 1 \ldots n \}} \left ( \hat{p}^{(\SC)}(\hat{y} \,|\, x) - \mathbb{I}[\hat{y} = y] \right )^2, \\
            \mathcal{E}_{\text{Model}}^{(\PP)} &= \sum_{\hat{t}\in\mathcal{R}} \left ( \hat{p}^{(\PP)}(\hat{t} \,|\, x) - \mathbb{I}[g(\hat{t}) = y] \right )^2.             
        \end{cases}
    \end{equation}
    where $\mathcal{R} = \text{Set}(\tilde{t}_1, \ldots, \tilde{t}_n)$ is the set of reasoning paths sampled from the LLM.
    We can compute the difference between the model error of \SC and \PP as follows:
    \begin{equation}
        \begin{aligned}
            \mathcal{E}_{\text{Model}}^{(\SC)} - \mathcal{E}_{\text{Model}}^{(\PP)}
            &=  \sum_{\hat{y} \in \{g(\tilde{t}_i) \, | \, i = 1 \ldots n \}} \left ( \hat{p}^{(\SC)}(\hat{y} \,|\, x) - \mathbb{I}[\hat{y} = y] \right )^2 
             %- \sum_{i=1}^n \left ( \hat{p}^{(\PP)}(\tilde{t}_i \,|\, x) - \mathbb{I}[g(\tilde{t}_i) = y] \right )^2 \\
             - \sum_{\hat{t}\in\mathcal{R}} \left ( \hat{p}^{(\PP)}(\hat{t} \,|\, x) - \mathbb{I}[g(\hat{t}) = y] \right )^2 \\            
            &= \sum_{\hat{y} \in \{g(\hat{t}_i) \, | \, i = 1 \ldots n \}} 
            \underbrace{\left [ \left ( \hat{p}^{(\SC)}(\hat{y} \,|\, x) - \mathbb{I}[\hat{y} = y] \right )^2 - \sum_{\hat{t} \in \mathcal{R}} \mathbb{I}[g(\hat{t}) = \hat{y}] \left ( \hat{p}^{(\PP)}(\hat{t} \,|\, x) - \mathbb{I}[\hat{y} = y] \right )^2 \right ]}_{(A)} \\
        \end{aligned}
    \end{equation}
    Assuming infinite samplings, the unbiasedness of \SC gives us: 
    \begin{equation}
        \begin{aligned}
            \hat{p}^{(\SC)}(\hat{y} \,|\, x) 
            = \frac{1}{n} \sum_{i=1}^n \mathbb{I}[g(\tilde{t}_i) = \hat{y}] 
             = \sum_{\hat{t} \in \mathcal{R}} \mathbb{I}[g(\hat{t}) = \hat{y}] \cdot \hat{p}^{(\PP)}(\hat{t} \,|\, x).
        \end{aligned}
    \end{equation}
    For any $\hat{y}$, consider two cases: 
    \begin{enumerate}
        \item[(i)] If $\hat{y} = y$, then $\hat{y}$ is the correct answer. We have 
        \begin{equation}
            \begin{aligned}
                (A) &= \left ( \hat{p}^{(\SC)}(\hat{y} \,|\, x) - 1 \right )^2 
                % - \sum_{i=1}^n \mathbb{I}[g(\tilde{t}_i) = \hat{y}] \left ( \hat{p}^{(\PP)}(\tilde{t}_i \,|\, x) - 1 \right )^2 
                - \sum_{\hat{t} \in \mathcal{R}} \mathbb{I}[g(\hat{t}) = \hat{y}] \left ( \hat{p}^{(\PP)}(\hat{t} \,|\, x) - 1 \right )^2 \\
                &= \left ( \hat{p}^{(\SC)}(\hat{y} \,|\, x)^2 + 1 - 2 \hat{p}^{(\SC)}(\hat{y} \,|\, x) \right ) 
                % - \sum_{i=1}^n \mathbb{I}[g(\tilde{t}_i) = \hat{y}] \left ( \hat{p}^{(\PP)}(\tilde{t}_i \,|\, x)^2 + 1 - 2 \hat{p}^{(\PP)}(\tilde{t}_i \,|\, x) \right ) \\
                - \sum_{\hat{t} \in \mathcal{R}} \mathbb{I}[g(\hat{t}) = \hat{y}] \left ( \hat{p}^{(\PP)}(\hat{t} \,|\, x)^2 + 1 - 2 \hat{p}^{(\PP)}(\hat{t} \,|\, x) \right ) \\
                &= \hat{p}^{(\SC)}(\hat{y} \,|\, x)^2 + 1  
                % - \sum_{i=1}^n \mathbb{I}[g(\tilde{t}) = \hat{y}] \cdot \hat{p}^{(\PP)}(\tilde{t}_i \,|\, x)^2 
                -\sum_{\hat{t} \in \mathcal{R}} \mathbb{I}[g(\hat{t}) = \hat{y}] \cdot \hat{p}^{(\PP)}(\hat{t} \,|\, x)^2
                % - \sum_{i=1}^n \mathbb{I}[g(\tilde{t}) = \hat{y}] \\
                - \sum_{\hat{t} \in \mathcal{R}} \mathbb{I}[g(\hat{t}) = \hat{y}] \\
                &\leq \hat{p}^{(\SC)}(\hat{y} \,|\, x)^2 + 1  - \frac{\hat{p}^{(\SC)}(\hat{y} \,|\, x)^2}{\sum_{\hat{t} \in \mathcal{R}} \mathbb{I}[g(\hat{t}) = \hat{y}]} 
                % - \sum_{i=1}^n \mathbb{I}[g(\tilde{t}) = \hat{y}] \\
                - \sum_{\hat{t} \in \mathcal{R}} \mathbb{I}[g(\hat{t}) = \hat{y}] \\
            \end{aligned}
        \end{equation}
        Let $\hat{p}^{(\SC)}(\hat{y} \,|\, x)^2 = B^2$ and $\sum_{\hat{t} \in \mathcal{R}} \mathbb{I}[g(\hat{t}) = \hat{y}]=C$, then
        \begin{equation}
            \begin{aligned}
                (A) &\leq B^2 + 1 - \frac{B^2}{C} - C \\
                &=\frac{1}{C}\left [  B^2C + C - B^2 - C^2 \right ] \\
                &=\frac{1}{C}\left [  (C-B^2)(1-C) \right ] \\
                &\leq 0.
            \end{aligned}
        \end{equation}
        Equality holds if $\sum_{\hat{t} \in \mathcal{R}} \mathbb{I}[g(\hat{t}) = \hat{y}]= C = 1$. This indicates that $(A)<0$ if the consistency function is effective at least once, making $\sum_{\hat{t} \in \mathcal{R}} \mathbb{I}[g(\hat{t}) = \hat{y}]= C >1$.
        \item[(ii)] If $\hat{y} \neq y$, then $\hat{y}$ is incorrect. Assuming distinct answers for wrong reasoning paths, we have $\sum_{\hat{t} \in \mathcal{R}} \mathbb{I}[g(\hat{t}) = \hat{y}] = 1$, thus
        \begin{eqnarray}
            \begin{aligned}
                (A) &= \left [ \left ( \hat{p}^{(\SC)}(\hat{y} \,|\, x) - \mathbb{I}[\hat{y} = y] \right )^2 
                - \sum_{\hat{t} \in \mathcal{R}} \mathbb{I}[g(\hat{t}) = \hat{y}] \left ( \hat{p}^{(\PP)}(\hat{t} \,|\, x) - \mathbb{I}[\hat{y} = y] \right )^2 \right ] \\
                &= \left [ \left ( \hat{p}^{(\SC)}(\hat{y} \,|\, x) 
                - \mathbb{I}[\hat{y} = y] \right )^2 - \left ( \hat{p}^{(\SC)}(\hat{y} \,|\, x) - \mathbb{I}[\hat{y} = y] \right )^2 \right ] = 0,
            \end{aligned}
        \end{eqnarray}
        since only one $\hat{t}\in \mathcal{R}$ satisfying that $g(\hat{t})$ equals the incorrect answer $\hat{y}$.
    \end{enumerate}
    Therefore, $\mathcal{E}_{\text{Model}}^{(\PP)} - \mathcal{E}_{\text{Model}}^{(\SC)} \leq 0$, indicating that the model error of self-consistency is always less than or equal to the model error of perplexity under our assumptions. Moreover, if the consistency function is effective at least once, the model error of self-consistency is strictly less than the model error of perplexity.
\end{proof}

\subsection{Proof of Theorem~\ref{thm:thm1}} \label{app:thm1}
\begin{proof}
For given answer $\hat{y}$, the estimated probability of \PC is $\hat{p}(\hat{y} \,|\, x) = \sum_{i=1}^n \I[g(\tilde{t}_i) = \hat{y}] p(\tilde{t}_i \,|\, x)$, allowing the reasoning error of \PC can be computed as follows.
    \begin{equation*}
        \begin{aligned}
          \mathcal{E}^{(\PC)}(\hat{p}) &= \E_{\tilde{t}_i \sim p(t \,|\, x)} \big[( \hat{p}(\hat{y} \,|\, x) - \I[\hat{y} = y] )^2 \big] \\
	& =\E_{\tilde{t}_i \sim p(t \,|\, x)}\big[(\hat{p}(\hat{y} \,|\, x) - p(\hat{y} \,|\, x) + p(\hat{y} \,|\, x) - \I[\hat{y} = y] )^2 \big].
        \end{aligned}
        \end{equation*}
We define $k := |\{\tilde{t} \mid g(\tilde{t}) = \hat{y}\}|$, which means that how many reasoning paths whose answers are $\hat{y}$ can be covered given limited sampling budget. 
Note that we further have $ \E_{\tilde{t} \sim p(t \,|\, x)} [\I[g(\tilde{t}) = \hat{y}] p(\tilde{t} \,|\, x)] = \frac{1}{k} {p}(\hat{y} \,|\, x)$, thus
\begin{equation*}
\begin{aligned}
\E_{\tilde{t}_i \sim p(t \,|\, x)} [\hat{p}(\hat{y} \,|\, x)] &= \E \big[\sum_{i=1}^n \I[g(\tilde{t}_i) = \hat{y}] p(\tilde{t}_i \,|\, x) \big] = \Big(1 - \big(1 - \frac{1}{k} p(\hat{y} \,|\, x)\big)^n \Big) p(\hat{y} \,|\, x) = (1-\alpha^n)p(\hat{y} \,|\, x),
\end{aligned}
\end{equation*}
where $\alpha := \frac{1}{k} p(\hat{y} \,|\, x)$.
Again, we have
\begin{equation*}
\begin{aligned}
& \E_{\tilde{t}_i \sim p(t \,|\, x)}[\hat{p}(\hat{y} \,|\, x) - p(\hat{y} \,|\, x)] = - \big(1 - \alpha\big)^n p(\hat{y} \,|\, x))  \\
& \E_{\tilde{t}_i \sim p(t \,|\, x)}[(\hat{p}(\hat{y} \,|\, x) - p(\hat{y} \,|\, x))^2] = \Big(1 - \big(1 - \alpha\big)^n \Big) \big(1 - \alpha\big)^n p(\hat{y} \,|\, x))^2   
\end{aligned}
\end{equation*}

Hence,
\begin{equation*}
    \begin{aligned}
      \mathcal{E}^{(\PC)}(\hat{p}) &= \E_{\tilde{t}_i \sim p(t \,|\, x)} \big[( \hat{p}(\hat{y} \,|\, x) - \I[\hat{y} = y] )^2 \big] \\
	& =\E_{\tilde{t}_i \sim p(t \,|\, x)}\big[(\hat{p}(\hat{y} \,|\, x) - p(\hat{y} \,|\, x) + p(\hat{y} \,|\, x) - \I[\hat{y} = y] )^2 \big] \\
	&= (1 - \alpha)^{n} p(\hat{y} \,|\, x) \big(2\I[\hat{y}=y] - (1 + (1 - \alpha)^{n}) p(\hat{y} \,|\, x) \big) +  \left( p(\hat{y} \,|\, x) - \I[\hat{y}_i = y] \right )^2,
    \end{aligned}
    \end{equation*}
which completes the proof.
\end{proof}

\subsection{Proof of Theroem~\ref{thm:thm2}} \label{app:thm2}
\begin{proof}
Let us set the pruning threshold by $\tau := p(y \,|\, x)$. Then, we have
\begin{equation*}
    \begin{aligned}
      \mathcal{E}^{(\RPC)}(\hat{p}) 
      = & \underbrace{ \alpha p(\hat{y} \,|\, x) \big(2\I[\hat{y}=y] - (1 + \alpha) p(\hat{y} \,|\, x) \big) \I[(p(\hat{y}) \,|\, x) < \tau]}_{\text{Estimation Error}} \\
        & \qquad + \underbrace{\left( p(\hat{y} \,|\, x) - \I[\hat{y}_i = y] \right )^2  \I[(p(\hat{y}) \,|\, x) < \tau]}_{\text{Model Error}} \\
    \end{aligned}
\end{equation*}
However, we only have an estimation of $p(\hat{y} \,|\, x)$, i.e., $\hat{p}(\hat{y} \,|\, x) = k \E_{\tilde{t} \sim p(t \,|\, x)} [\I[g(\tilde{t}) = \hat{y}] p(\tilde{t} \,|\, x)] \approx \frac{k}{\hat{k}} \sum_{i=1}^k p(\tilde{t}_i \,|\, x)$, where $\tilde{t}_1, \dots, \tilde{t}_{\hat{k}}$ are $\hat{k}$ sampled reasoning paths whose answer is $\hat{y}$.
Therefore, the reasoning error of our approximate version can be computed by
\begin{equation*}
    \begin{aligned}
      \hat{\mathcal{E}}^{(\RPC)}(\hat{p}) 
      = & \underbrace{ \alpha(p) p(\hat{y} \,|\, x) \big(2\I[\hat{y}=y] - (1 + \alpha(p)) p(\hat{y} \,|\, x) \big) \I[\frac{1}{k} \sum_{i=1}^{\hat{k}} p(\tilde{t}_i \,|\, x) < \frac{1}{m}\tau]}_{\text{Estimation Error}} \\
        & \qquad + \underbrace{\left( p(\hat{y} \,|\, x) - \I[\hat{y}_i = y] \right )^2  \I\big[\frac{1}{\hat{k}} \sum_{i=1}^{\hat{k}} p(\tilde{t}_i \,|\, x) < \frac{1}{m}\tau\bigskip
        ]}_{\text{Model Error}} \\
    \end{aligned}
\end{equation*}

Hence, we only need to consider the probability $\frac{1}{\hat{k}} \sum_{i=1}^{\hat{k}} p(\tilde{t}_i \,|\, x) > \frac{1}{m} \tau$.
Using Hoeffding's inequality, we can obtain that
\begin{equation*}
\mathbb{P}\big(\frac{1}{\hat{k}} \sum_{i=1}^{\hat{k}} p(\tilde{t}_i \,|\, x) - \frac{1}{k} p(\hat{y} \,|\, x) \ge \tau \big)  \leq \exp\Big(-\frac{2\hat{k}\gamma^2}{ p(\hat{y} \,|\, x)^2}\Big)
\end{equation*}
We set $\gamma = \tau - \frac{1}{k} p(\hat{y} \,|\, x) = \tau + \alpha - 1$, then 
\begin{equation*}
\mathbb{P}\big(\frac{1}{\hat{k}} \sum_{i=1}^{\hat{k}} p(\tilde{t}_i \,|\, x) \ge \tau \big)  \leq \exp\Big(-{2\hat{k}k^2} (1 - \frac{\tau}{1 - \alpha})^2\Big).
\end{equation*}
Hence, we complete the proof.
\end{proof}

\lstset{
  language=python,
  basicstyle=\ttfamily\small,
  keywordstyle=\color{blue},
  stringstyle=\color{red},
  commentstyle=\color{gray},
  morecomment=[l][\color{magenta}]{\#},
}

\section{Pseudo Code of \RPC Method}
\label{sec:appendix-rpc}

In this section, we provide the pseudo code of \RPC. 
The output of Algorithm~\ref{alg:rpc} is a set of reasoning paths with the highest confidence.
The extraction function $g(\cdot)$ can be used to transform the reasoning paths to answers.

\renewcommand{\algorithmicrequire}{\textbf{Input:}}
\renewcommand{\algorithmicensure}{\textbf{Output:}}
\begin{algorithm}[ht]
    \caption{Reasoning-pruning Perplexity Consistency}
    \label{alg:rpc}
    \begin{algorithmic}[1]
    \REQUIRE 
        \STATE Sampled Reasoning paths $\tilde{t}_1, \ldots, \tilde{t}_n$ 
        \STATE LLM Internal Probabilities $p_1, \ldots, p_n$
        \STATE Consisitency function $\mathbb{I}_C(\cdot, \cdot)$
    \ENSURE Most-confident reasoning paths with probabilities
    
    \STATE \COMMENT{\textbf{\underline{Phase 1}: Reasoning Pruning}}
    \STATE $(k_1, \lambda_1, k_2, \lambda_2, w_1, w_2) \gets$ Model probability distribution parameters from $p_1, \ldots, p_n$ \COMMENT{Using \autoref{eq:weibull-mix}}
    \STATE $p_{\text{mean}} \gets \frac{1}{n}\sum_{i=1}^n p_i$
    \STATE $\mathrm{I}_{\text{retain}} \gets \{i \mid P_{\text{High}}(p_i) > 0.5 \text{ or } p_i \geq p_{\text{mean}}\}$ \COMMENT{$P_{\text{High}}$ is defined in \autoref{eq:weibull-prob}}
    \STATE \COMMENT{\textbf{\underline{Phase 2}: Perplexity Consistency}}
    \STATE $\mathrm{U} \gets \text{Set}(\tilde{t}_1, \ldots, \tilde{t}_n)$ 
    \STATE $\mathrm{I}_{\text{unique}} \gets \{i \mid \tilde{t}_i \in \mathrm{U} \text{ and } i \in \mathrm{I}_{\text{retain}}\}$ 
    \FOR{each reasoning path $\tilde{t} \in \mathrm{U}$}
        \STATE $\mathrm{C}(\tilde{t}) \gets \sum_{i \in \mathcal{I}_{\text{retain}}} \mathbb{I}_C[\tilde{t}, \tilde{t}_i] p_i$
    \ENDFOR

    \STATE $\text{C}_{\text{max}} \gets \max_{\tilde{t} \in \mathrm{U}} \mathrm{C}(\tilde{t})$
    \STATE {\bfseries return} $\{(\tilde{t}, \mathrm{C}(\tilde{t})) \mid \tilde{t} \in \mathrm{U}, \text{C}(\tilde{t}) = \mathrm{C}_{\text{max}}\}$
 \end{algorithmic}
\end{algorithm}

% \section{Details of \PC and \RPC methods}

% \algrenewcommand\algorithmicrequire{\textbf{Input:}}
% \algrenewcommand\algorithmicensure{\textbf{Output:}}

% \begin{algorithm}
% \caption{The pseudo-code of \PC approach. }\label{alg:PC}
% \begin{algorithmic}
% \Require Response and corresponding probability pairs $ \mathbf{R} = \left \{ (\mathrm{resp}_i, \mathrm{prob}_i) \right \}_{i=1}^n$ of reasoning paths; \\
% ~~~~~~~~~Consisitency function $\mathbb{I}_C(\cdot, \cdot)$; Answer extraction function $G(\cdot)$.
% \Ensure Answer and corresponding confidence pairs $\{ (\mathrm{ans}_i, \mathrm{conf}_i) \}_{i=1}^m$. 
% \State $\mathbf{ans} \gets \left \{G(\mathrm{resp}_i) ~|~ (\mathrm{resp}_i, \mathrm{prob}_i) \in \mathbf{R} \right \}$
% \State $m \gets |\mathbf{ans}|$
% \State $\mathbf{conf} \gets $ A $m$-sized float array initialized by 0
% \State $\mathbf{R}^{\star} \gets \mathrm{Unique}(\mathbf{R})$ 
% \For{$i = 1 \ldots |\mathbf{R}^{\star}|$}

% \State $k \gets$ Find 
% \EndFor
% % \For{$i=1 \ldots n$}
% % \State 
% % \EndFor
% \State $y \gets 1$
% \State $X \gets x$
% \State $N \gets n$
% \While{$N \neq 0$}
% \If{$N$ is even}
%     \State $X \gets X \times X$
%     \State $N \gets \frac{N}{2}$  \Comment{This is a comment}
% \ElsIf{$N$ is odd}
%     \State $y \gets y \times X$
%     \State $N \gets N - 1$
% \EndIf
% \EndWhile
% \end{algorithmic}
% \end{algorithm}

\section{Detailed Experimental Settings}
\label{app:exp-details}

\subsection{Datasets}
\label{sec:datasets}
For mathematical reasoning tasks, we evaluate our proposed methods and comparison methods on four mathematical datasets that include MATH, MathOdyssey, OlympiadBench, and AIME datasets. 
\begin{itemize}
    \item MATH dataset~\citep{hendrycks2021math} is a dataset comprised of challenging competition math problems and we use its 5,000 testing data for evaluation. 
    \item MathOdyssey dataset~\citep{Fang24Odyssey} contains 387 problems, covering advanced high-school level, university-level, and Olympiad-level mathematics. 
    \item OlympiadBench dataset~\citep{He24OlympiadBench} contains 8,476 Olympiad-level mathematics and physics problems. We select the English problems without images, resulting in a testing dataset of 1,284 problems. 
    \item AIME dataset~\citep{AIME} contains 993 test problems collected from the American Invitational Mathematics Examination, spanning from 1983 to 2024.
\end{itemize}

For code generation tasks, we conduct experiments on three common benchmark datasets. 
HumanEval~\citep{Codex2021} contains 164 hand-written Python programming problems. 
MBPP~\citep{MBPP2021}(sanitized version) consists of 427 entry-level programming problems. 
We also include the introductory-level problems of APPS~\citep{APPS2021}, which contains 1000 problems.

\subsection{Detailes of Mathematical Reasoning Task}

For all the experiments in the main paper, we use a sampling temperature of 1.0 and set the top-k parameter to 0.95. 

\paragraph{Prompt for Math Reasoning Tasks.}

The InternLM2-MATH-Plus 1.8B and 7B models are chat models that facilitate conversations between two roles: ``user'' and ``assistant''. 
The prompt for the “user” role is provided in Prompt~\ref{pt:internlm-math}. Similarly, the prompt for the DeepSeek-Math 7B model is shown in Prompt~\ref{pt:deepseek-math}.

\begin{figure}[htb!]
    \begin{promptbox}[label=pt:internlm-math]{Prompt for InternLM-2-Math-Plus}
    Problem:\textbackslash n\{instruction\}\textbackslash n Let's think step by step\textbackslash n Solution:\textbackslash n
    \end{promptbox}
\end{figure}

\begin{figure}[htb!]
    \begin{promptbox}[label=pt:deepseek-math]{Prompt for DeepSeek-Math}
    {instruction}\textbackslash n Please reason step by step, and put your final answer within \textbackslash\textbackslash boxed\{\{\}\}.
    \end{promptbox}
\end{figure}

\paragraph{Prompt for Math Verbalized Method.}

We observed that the tuned math models are challenging to prompt for generating confidence. Therefore, we adopted the methods from \citet{tian2023just} to calculate the probability based on the likelihood of the first generated ``True'' token and the first generated ``False'' token. The corresponding prompt is provided in Prompt~\ref{pt:math-verb}.

\begin{figure}[htb!]
    \begin{promptbox}[label=pt:math-verb]{Prompt for DeepSeek-Math}
    Question: {question}\textbackslash n Proposed Answer: {answer}\textbackslash n Is the proposed answer:\textbackslash n \textbackslash t(A) True or\textbackslash n \textbackslash t(B) False?\textbackslash n The proposed answer is:
    \end{promptbox}
\end{figure}

\subsection{Detailes of Code Generation Task}

\paragraph{Code Generation.} 
On the code generation task, we let LLM generate a code snippet to solve a given programming problem, and then evaluate its functional correctness based on the ground-truth test cases provided by the dataset. In detail, we set the top \textit{p} to 0.95, and the max generation length to 1024. For code snippet post-processing, we first extract the code text from the code block surrounded by triple-backticks(\texttt{```}), and then we follow~\citet{Codex2021} to truncate the generated code snippet before the following stop sequences: ``\textbackslash nclass", ``\textbackslash ndef", ``\textbackslash n\#", ``\textbackslash nif", ``\textbackslash nprint". At the same time, we also obtain the log-probability of each token from the LLM response. For "verbalization" setting, the verbalized confidence is also extracted from the text generated by LLM along with the code snippet. 

\paragraph{Self-consistency on Code.}
We follow \citet{chen2022codet} to sample 100 test cases for each programming problem from the same model. Then, we achieved self-consistency in code at the semantic equivalence level, which is based on the execution behavior of any two codes on this set of test cases. More formally, we implemented the consistency function $\mathbb{I}_C(\cdot,\cdot)$ as an indicator function that indicates whether two codes are semantically equivalent, i.e., $\mathbb{I}_C(x,y) = 1$ if and only if code $x$ and $y$ execute the same result on this set of test cases.

\paragraph{Prompt for Generating Code.}
The prompt for generating code consists of a header, a functional signature, and a docstring and LLM needs to implement the body of this function. An Illustration is shown in Prompt~\ref{pt:generate-code}.

\begin{figure}[ht]
\begin{promptbox}[label=pt:generate-code]{Prompt for Generating Code}
\begin{lstlisting}
from typing import List
def has_close_elements(numbers: List[float], 
        threshold: float) -> bool:
""" Check if in given list of numbers, 
are any two numbers closer to
each other than given threshold.
"""\end{lstlisting}
\end{promptbox}
\end{figure}

\paragraph{Prompt for Generating Test Cases.}
For generating test cases, we implemented the function body with a ``pass'' statement on the basis of the prompt to generate the code, and added a comment to require the LLM to generate test cases for the programming problem. An illustration is shown in Prompt~\ref{pt:generate-cases}.

\begin{figure}[ht]
\begin{promptbox}[label=pt:generate-cases]{Prompt for Generating Test Cases.}
\begin{lstlisting}
from typing import List
def has_close_elements(numbers: List[float], 
        threshold: float) -> bool:
""" Check if in given list of numbers, are any two numbers closer to
each other than given threshold.
"""
pass
# check the correctness of has_close_elements
assert
\end{lstlisting}
\end{promptbox}
\end{figure}

\paragraph{Prompt for Code Verbalized Method.}
For generating code with verbalized confidence, we added instructions for generating verbalized confidence, as well as format requirements to facilitate the extraction of code and confidence score. We also gave a simple example to help LLM understand the format requirements at the end of the prompt. 
An Illustration is shown in Prompt~\ref{pt:code-verbalized}.

\begin{figure}[ht]
\begin{promptbox}[label=pt:code-verbalized]{Prompt for Code Verbalized Method}
Come up with a solution that solves the following programming question and
provide your confidence score in this solution like 0\%, 10\%, ... 100\%.
\begin{lstlisting}
import heapq as hq
def heap_queue_largest(nums,n):
'''
Write a function to find the n largest integers from a given list 
of numbers, returned in descending order.
'''
```
\end{lstlisting}
Format requirement: output in the form of the following example. Do not
provide any additional explanations.
Here is an output example:
Solution:
\begin{lstlisting}
```python
your code ...
```
\end{lstlisting}
Confidence: 
\end{promptbox}
\end{figure}
\section{Detailed Experimental Results}
\label{app:detailed-results}

\subsection{Performance on GSM8k dataset.} 
In Section~\ref{sec:problem}, we analyze the GSM8k dataset, which is relatively easy that allows accurate estimation of ground-truth probabilities with 100 samples. Due to this characteristic, we exclude the GSM8k dataset from our main experiments, as a limited number of samples is sufficient for accurate confidence evaluation.
\autoref{tab:gsm8k} shows the performance of InternLM-2-MATH-Plus 7B model on GSM8k with various sampling temperatures, where the number of samples is set to $n=6$ to maintain consistency with \autoref{fig:motivation-estimation-error}. The results show that the \RPC method outperforms comparison methods.

\begin{table}[ht]
    \centering
    \caption{Performance on GSM8k datasets using InternLM-2-MATH-Plus 7B model with diverse sampling temperatures when sampling size $n=6$. The best performance is highlighted in \textbf{bold}. The results show that the \RPC approach outperforms the \SC method. }
    \vskip 0.15in
    \begin{center}
    \begin{small}
    \begin{sc}
    \label{tab:gsm8k}
    \begin{tabular}{l|cccc}
    \bottomrule
    \toprule
     & \PP & \Verb & \SC & \RPC  \\
    \midrule
    T=1.0 & 86.97 $\pm$ 0.29 & 63.67 $\pm$ 0.98 & 89.32 $\pm$ 0.26 & \textbf{89.45 $\pm$ 0.38} \\
    T=1.1 & 86.78 $\pm$ 0.43 & 62.25 $\pm$ 1.00 & 89.44 $\pm$ 0.35 & \textbf{89.51 $\pm$ 0.36} \\
    T=1.3 & 86.65 $\pm$ 0.57 & 61.29 $\pm$ 0.84 & 88.92 $\pm$ 0.32 & \textbf{89.08 $\pm$ 0.57} \\
    \bottomrule
    \toprule
    \end{tabular}
    \end{sc}
    \end{small}
    \end{center}
    \vskip -0.1in
\end{table}

\subsection{Results with High Sampling Temperature. }

Using a high sampling temperature enables language models to produce more diverse outputs, potentially enhancing reasoning performance. 
However, it also leads to an increase in estimation error. 
To investigate the effectiveness of our approaches in addressing the estimation error issue, we conducted experiments with higher sampling temperatures (T = 1.1 and T = 1.3) using the InternLM-2-MATH-Plus 7B model. The results in \autoref{tab:temp-performance} indicate that our \RPC approach consistently surpasses baseline methods. Notably, a significant performance gap persists between \RPC and \SC, indicating that our methods effectively tackle the estimation error issue even under high-temperature sampling conditions.

\begin{table}[t]
    \centering
    \caption{Performance Comparison of different models and different parameter scales. The accuracy is reported as the mean and stdev. The best performance is highlighted in \textbf{bold}. The results show that our \RPC approach consistently outperforms existing methods.}
    \label{tab:temp-performance}
    % \vskip 0.15in
    \begin{center}
    \begin{small}
    \begin{sc}
    % \resizebox{\linewidth}{!}{
    \begin{tabular}{l|cccc}
    \bottomrule
    \toprule
    \multirow{2}{*}{Method} & \multicolumn{4}{c}{\textbf{ Temperature = 1.1}} \\
    \cmidrule(lr){2-5}      & MATH & MathOdyssey & OlympiadBench & AIME \\
    \midrule
    \PP & 47.35 $\pm$ 0.16 & 28.59 $\pm$ 1.30 & ~~7.27 $\pm$ 0.23 & 6.02 $\pm$ 0.34 \\
    \Verb & 25.51 $\pm$ 0.23 & ~~9.41 $\pm$ 0.44 & ~~3.66 $\pm$ 0.16 & 3.07 $\pm$ 0.15 \\
    \SC & 50.66 $\pm$ 0.22 & 27.89 $\pm$ 0.43 & 10.74 $\pm$ 0.15 & 8.73 $\pm$ 0.24 \\
    \hline
    \rowcolor{gray!20} \RPC & \textbf{52.58 $\pm$ 0.14} & \textbf{32.98 $\pm$ 0.69} & \textbf{11.00 $\pm$ 0.24} & \textbf{9.30 $\pm$ 0.29} \\
    \bottomrule
    \toprule
    \multirow{2}{*}{Method} & \multicolumn{4}{c}{\textbf{Temperature = 1.3}} \\
    \cmidrule(lr){2-5}      & MATH & MathOdyssey & OlympiadBench & AIME \\
    \midrule
    \PP & 47.58 $\pm$ 0.31 & 26.38 $\pm$ 1.41 & ~~7.76 $\pm$ 0.46 & 6.50 $\pm$ 0.41 \\
    \Verb & 24.62 $\pm$ 0.33 & ~~8.60 $\pm$ 0.26 & ~~3.11 $\pm$ 0.17 & 2.29 $\pm$ 0.12 \\
    \SC & 50.65 $\pm$ 0.14 & 27.61 $\pm$ 0.67 & 10.49 $\pm$ 0.18 & 8.02 $\pm$ 0.20 \\
    \hline
    \rowcolor{gray!20} \RPC & \textbf{53.12 $\pm$ 0.14} & \textbf{33.19 $\pm$ 0.56} & \textbf{10.91 $\pm$ 0.18} & \textbf{8.83 $\pm$ 0.23} \\
    \bottomrule
    \toprule
    \end{tabular}
    \end{sc}
    \end{small}
    \end{center}
    \vskip -0.2in
\end{table}

\subsection{Performance with Diverse Number of Samplings}

In the \autoref{fig:InternLM2-7B-Accuracy}, we plot the accuracy of the InternLM-2-MATH-Plus 7B model on four mathematical reasoning datasets with different sample sizes $n$. Here, we give the detailed results on different models and different sampling temperatures.

\begin{figure}[ht]
    \vskip 0.2in
    \begin{center}
        \begin{minipage}[t]{\textwidth}
            \centering
            \begin{subfigure}[t]{0.23\textwidth}
                \centering
        \includegraphics[width=\linewidth]{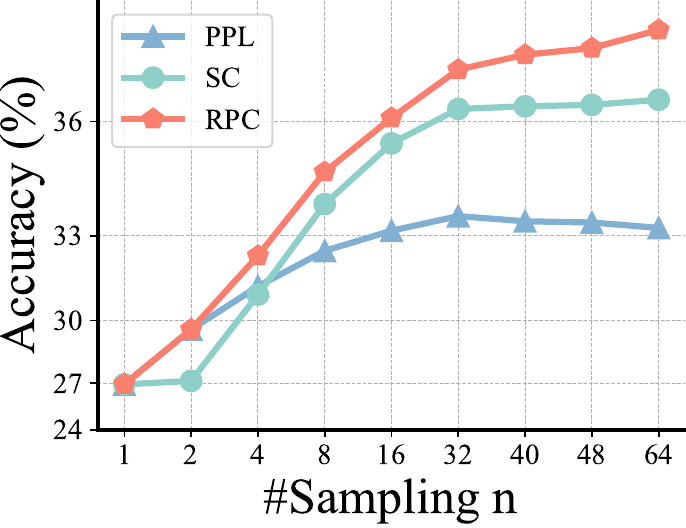}
                \vskip -0.3em
                \caption{MATH}
                \label{fig:MATH-Accuracy}
            \end{subfigure}
            \hfill
            \begin{subfigure}[t]{0.23\textwidth}
                \centering
                \includegraphics[width=\linewidth]{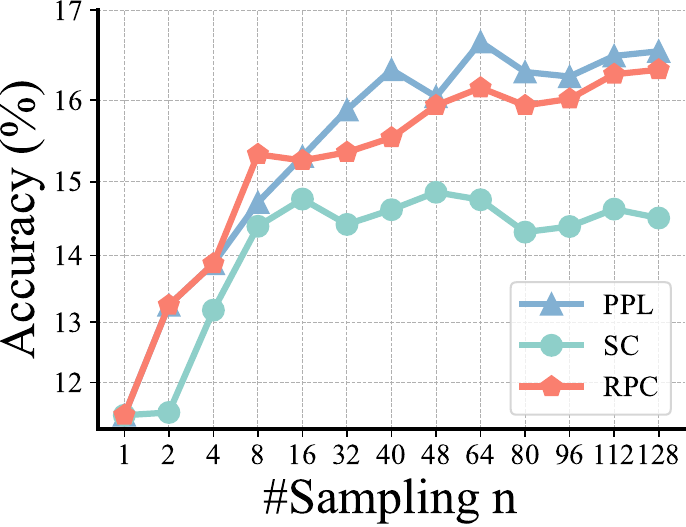}
                \vskip -0.3em
                \caption{MathOdyssey}
                \label{fig:MathOdyssey-Accuracy}
            \end{subfigure}
            \hfill
            \begin{subfigure}[t]{0.23\textwidth}
                \centering
                \includegraphics[width=\linewidth]{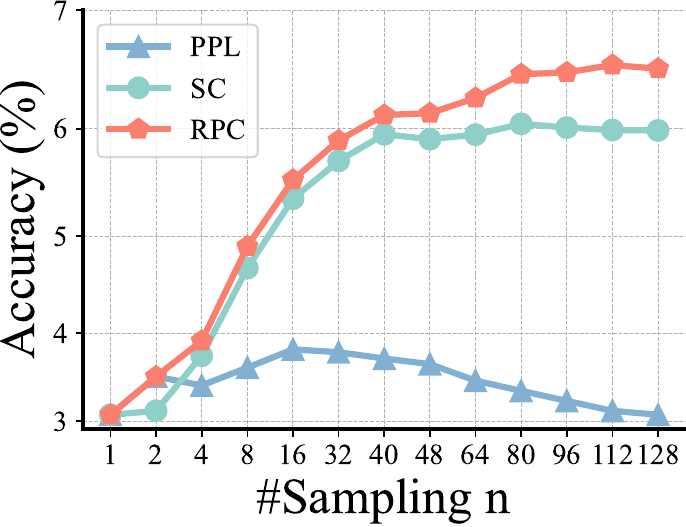}
                \vskip -0.3em
                \caption{OlympiadBench}
            \end{subfigure}
            \hfill
            \begin{subfigure}[t]{0.23\textwidth}
                \centering
                \includegraphics[width=\linewidth]{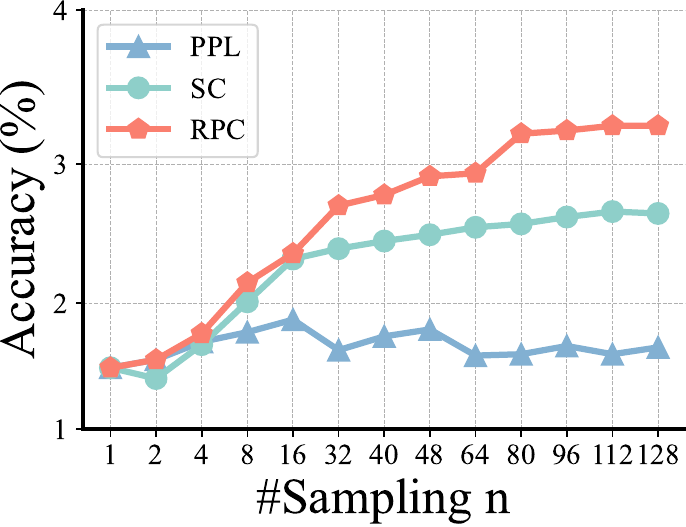}
                \vskip -0.3em
                \caption{AIME}
            \end{subfigure}
            \vskip -0.5em
            \caption{The accuracy of InternLM-2-MATH-Plus 1.8B model on four mathematical reasoning datasets with different sample sizes $n$.}
            \label{fig:InternLM2-1_8B-Accuracy}
        \end{minipage}
    \end{center}
    \vskip -0.2in
\end{figure}

\paragraph{Different Models Scales.} 
The performance of relateively small model, InternLM-2-MATH-Plus 1.8B, is presented in \autoref{fig:InternLM2-1_8B-Accuracy}.  Similar conclusions can be drawn from these results. For the MathOdyssey dataset, the \PP method shows superior performance compared to other methods, which can be attributed to the relatively low model error of \PP on this dataset, allowing the perplexity-based approach to function effectively. Furthermore, the \RPC method consistently outperforms the \SC method, which demonstrates its ability to enhance the convergence properties of the \SC method. 

\begin{figure}[ht]
    \vskip 0.2in
    \begin{center}
        \begin{minipage}[t]{\textwidth}
            \centering
            \begin{subfigure}[t]{0.23\textwidth}
                \centering
        \includegraphics[width=\linewidth]{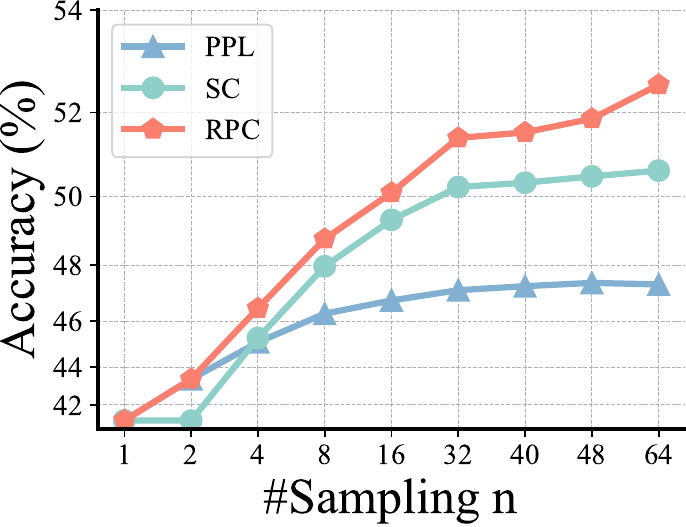}
                \vskip -0.3em
                \caption{MATH}
                \label{fig:MATH-Accuracy}
            \end{subfigure}
            \hfill
            \begin{subfigure}[t]{0.23\textwidth}
                \centering
                \includegraphics[width=\linewidth]{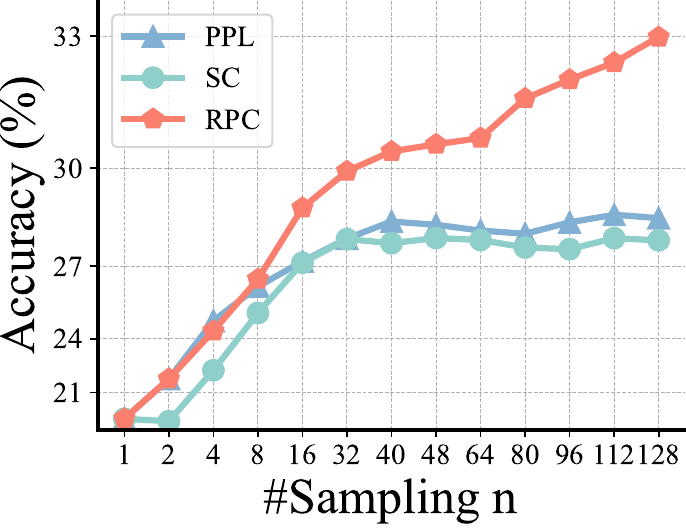}
                \vskip -0.3em
                \caption{MathOdyssey}
                \label{fig:MathOdyssey-Accuracy}
            \end{subfigure}
            \hfill
            \begin{subfigure}[t]{0.23\textwidth}
                \centering
                \includegraphics[width=\linewidth]{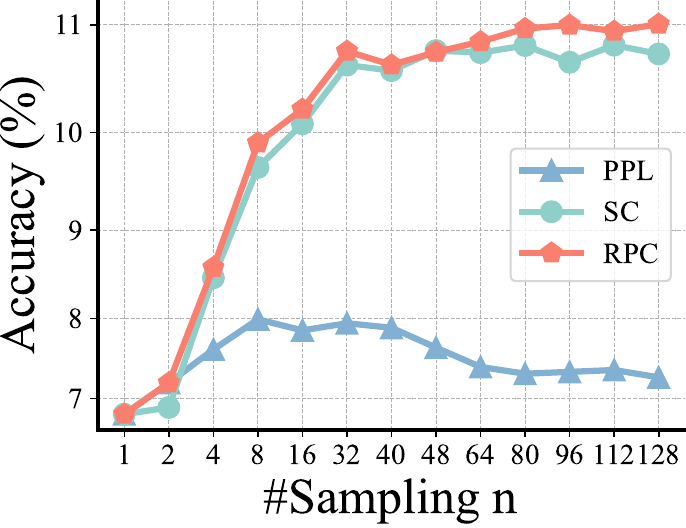}
                \vskip -0.3em
                \caption{OlympiadBench}
            \end{subfigure}
            \hfill
            \begin{subfigure}[t]{0.23\textwidth}
                \centering
                \includegraphics[width=\linewidth]{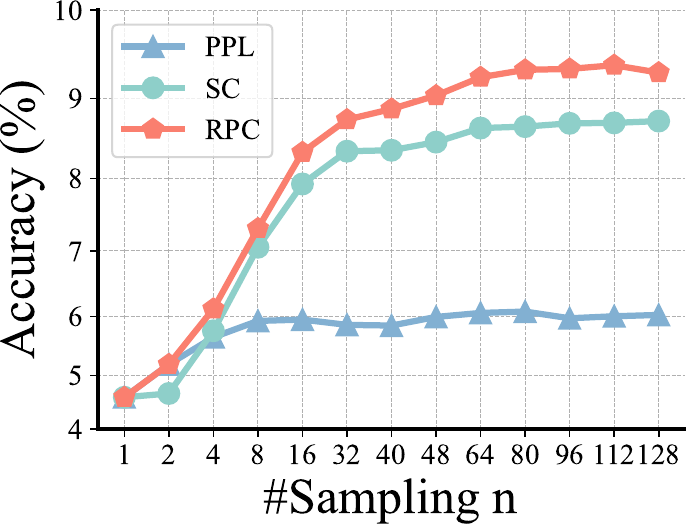}
                \vskip -0.3em
                \caption{AIME}
            \end{subfigure}
            \vskip -0.5em
            \caption{The accuracy of InternLM-2-MATH-Plus 7B model on four mathematical reasoning datasets with different sample sizes $n$. The sampling temperature is set to 1.1.}
            \label{fig:InternLM2-7B-Accuracy-T1.1}
        \end{minipage}
    \end{center}
    \vskip -0.2in
\end{figure}

\begin{figure}[ht]
    \vskip 0.2in
    \begin{center}
        \begin{minipage}[t]{\textwidth}
            \centering
            \begin{subfigure}[t]{0.23\textwidth}
                \centering
        \includegraphics[width=\linewidth]{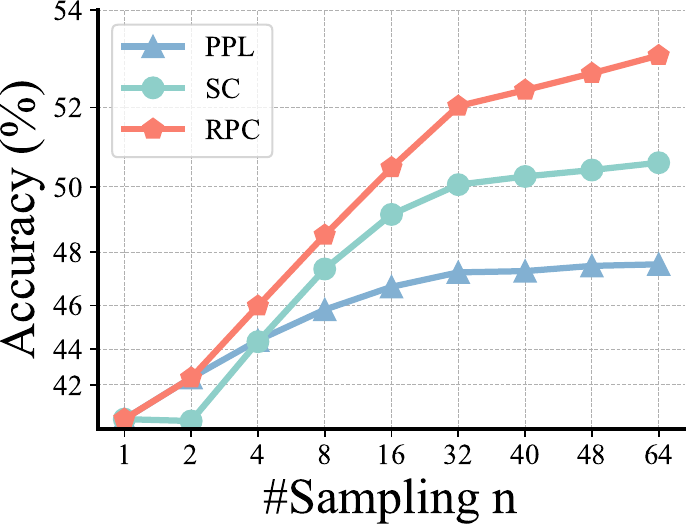}
                \vskip -0.3em
                \caption{MATH}
                \label{fig:MATH-Accuracy}
            \end{subfigure}
            \hfill
            \begin{subfigure}[t]{0.23\textwidth}
                \centering
                \includegraphics[width=\linewidth]{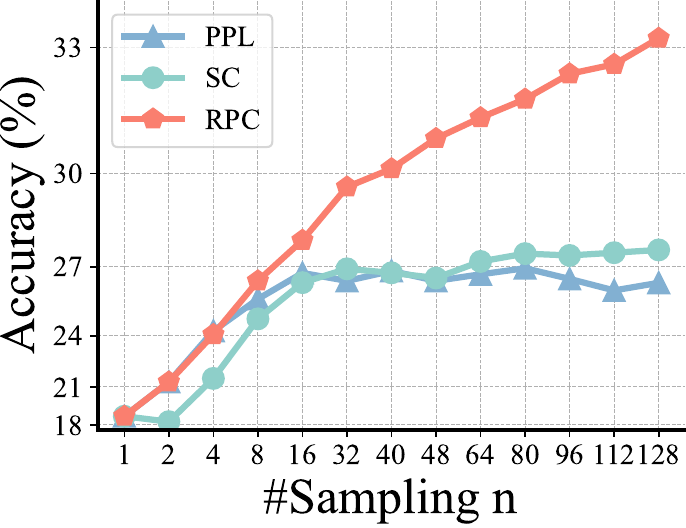}
                \vskip -0.3em
                \caption{MathOdyssey}
                \label{fig:MathOdyssey-Accuracy}
            \end{subfigure}
            \hfill
            \begin{subfigure}[t]{0.23\textwidth}
                \centering
                \includegraphics[width=\linewidth]{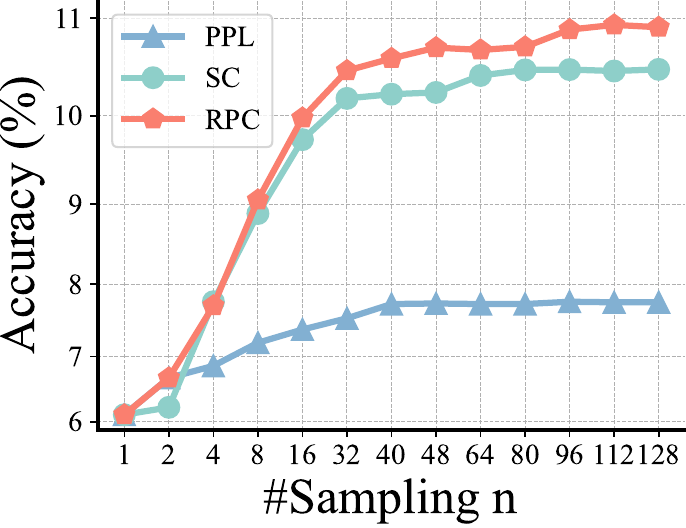}
                \vskip -0.3em
                \caption{OlympiadBench}
            \end{subfigure}
            \hfill
            \begin{subfigure}[t]{0.23\textwidth}
                \centering
                \includegraphics[width=\linewidth]{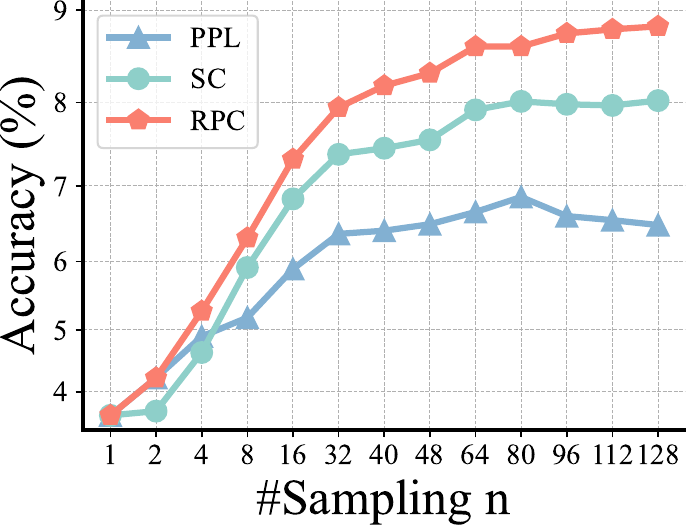}
                \vskip -0.3em
                \caption{AIME}
            \end{subfigure}
            \vskip -0.5em
            \caption{The accuracy of InternLM-2-MATH-Plus 7B model on four mathematical reasoning datasets with different sample sizes $n$. The sampling temperature is set to 1.3.}
            \label{fig:InternLM2-7B-Accuracy-T1.3}
        \end{minipage}
    \end{center}
    \vskip -0.2in
\end{figure}

\paragraph{Different Sampling Temperatures.} We also evaluate the InternLM-2-MATH-Plus 7B model with sampling temperatures $T=1.1$ and $T=1.3$. The results are presented in \autoref{fig:InternLM2-7B-Accuracy-T1.1} and \autoref{fig:InternLM2-7B-Accuracy-T1.3}. 
The results demonstrate that the \RPC method effectively improves the model's reasoning performance at higher temperatures, as it leverages the increased diversity in sampling to enhance self-consistency. In contrast, the \SC method's performance deteriorates due to increased estimation errors at higher temperatures.

%%%%%%%%%%%%%%%%%%%%%%%%%%%%%%%%%%%%%%%%%%%%%%%%%%%%%%%%%%%%%%%%%%%%%%%%%%%%%%%
%%%%%%%%%%%%%%%%%%%%%%%%%%%%%%%%%%%%%%%%%%%%%%%%%%%%%%%%%%%%%%%%%%%%%%%%%%%%%%%

\end{document}